\documentclass[letterpaper,10pt]{article}

\usepackage{probml}

\ShortHeadings{Anchor-Based Heteroscedastic Noise for PBO}
\usepackage{blindtext}
\usepackage{wrapfig}

\usepackage{bm,amssymb,amsmath}

\makeatletter
\def\th@plain{%
  \thm@notefont{}% same as heading font
  \itshape % body font
}
\def\th@definition{%
  \thm@notefont{}% same as heading font
  \normalfont % body font
}
\makeatother

\def\1{\bm{1}}

\def\vv{{\bm{v}}}

\DeclareMathAlphabet{\mathsfit}{\encodingdefault}{\sfdefault}{m}{sl}
\SetMathAlphabet{\mathsfit}{bold}{\encodingdefault}{\sfdefault}{bx}{n}

\newcommand{\R}{\mathbb{R}}

\DeclareMathOperator*{\argmax}{argmax}

\usepackage{mathtools} % amsmath with fixes and additions
\usepackage{booktabs} % commands to create good-looking tables
\usepackage{tikz} % nice language for creating drawings and diagrams
\usepackage[most]{tcolorbox}
\usepackage{amssymb}
\usepackage{graphicx}
\usepackage{cleveref}
\usepackage{algorithm}
\usepackage{algorithmic}
\usepackage{thm-restate}
\usepackage{subfigure}
\usepackage{enumitem}

\newcommand{\dataset}{{\cal D}}
\newcommand{\bsigma}{\boldsymbol{\Sigma}}
\newcommand{\bmu}{\boldsymbol{\mu}}
\newcommand{\var}{{\mathbb V}}
\newcommand{\mean}{{\mathbb E}}
\newcommand{\normal}{{\mathcal{N}}}
\newcommand\X{{\bf{X}}}
\newcommand\x{{\bf{x}}}
\newcommand\s{{\bf{s}}}
\newcommand\y{{\bf{y}}}
\newcommand\z{{\bf{z}}}

\newcommand\ff{{\bf{f}}}
\newcommand\cc{{\bf{c}}}

\newtheorem{assumption}[theorem]{Assumption}

\graphicspath{ {aistats2025/} }

\begin{document}

\title{
  Anchor-Based Heteroscedastic Noise for Preferential Bayesian Optimization
}

\author[1,2,$\dagger$]{Marshal Sinaga}
\author[1,2]{Julien Martinelli}
\author[1,2,3]{Samuel Kaski}

\affil[1]{ELLIS Institute Finland}
\affil[2]{Aalto University}
\affil[3]{University of Manchester}
\affil[$\dagger$]{Correspondence to \url{marshal.sinaga@aalto.fi}}

\maketitle

\begin{abstract}
Preferential Bayesian optimization (PBO) learns latent utilities from pairwise comparisons, but most existing methods assume homoscedastic comparison noise. This is inadequate in human-in-the-loop settings, where a user may compare some designs reliably and others only hesitantly. We propose a heteroscedastic noise model for PBO: before optimization, the user provides a small set of reliable examples, called \emph{anchors}, and a kernel density estimator (KDE) turns these anchors into an input-dependent map of user uncertainty. We incorporate this map into preferential GP surrogates and derive risk-averse acquisition functions that trade off utility and ease of comparison. We further show that a risk-adjusted variant of the popular expected utility of the best option (EUBO) preserves the one-step Bayes-optimality guarantee up to an additive constant, and that under an idealized i.i.d.\ anchor model the KDE estimator enjoys standard consistency and concentration rates. Experiments on synthetic problems and human-preference datasets show improved risk-adjusted performance and clarify how anchor placement affects the method.
\end{abstract}

\section{Introduction}\label{sec:intro}

Preferential Bayesian Optimization (PBO) learns a latent utility function from pairwise comparisons rather than direct scalar evaluations \citep{chu2005preference,gonzalez2017preferential}. It is therefore well suited to human-in-the-loop design problems, where users are typically better at choosing between two options than assigning an absolute score \citep{kahneman2013prospect}. Applications include visual design optimization \citep{koyama2020sequential}, recommender systems \citep{webbook}, and expert-guided molecular design \citep{sundin_human---loop_2022,nahal2024human}.

% A key but under-modeled aspect of human feedback is that comparison difficulty is often input-dependent. An expert may compare two familiar designs confidently yet struggle in unfamiliar regions of the search space. Existing PBO methods almost always use homoscedastic comparison noise, which treats all duels as equally reliable and leads to \emph{risk-neutral} query selection \citep{gonzalez2017preferential,takeno2023towards,pmlr-v206-astudillo23a,principledpbo}. When two candidate pairs have similar expected utility but very different levels of user uncertainty, ignoring this heterogeneity wastes queries on hard-to-judge comparisons.

A key but under-modeled aspect of human feedback is that comparison difficulty is often input-dependent. For example, in expert-guided molecular design, a chemist may confidently compare familiar small molecules but be less reliable on unfamiliar chemical families or materials classes. Prior PBO methods all assume homoscedastic comparison noise, which treats all duels as equally reliable and leads to risk-neutral query selection~\citep{gonzalez2017preferential,takeno2023towards,pmlr-v206-astudillo23a,principledpbo}. When two candidate pairs have similar expected utility but different levels of user uncertainty, ignoring this heterogeneity can waste queries on hard-to-judge comparisons.

A natural reaction is to borrow heteroscedastic noise models from standard BO \citep{griffiths2021achieving,hebo,makarova2021risk}. In preferential learning, however, binary comparisons provide only ordinal information about latent utility, so flexible utility and noise models are difficult to disentangle. This makes it natural to look for an additional source of information about \emph{where} user judgments are likely to be reliable.

In many expert-facing applications, this information can be elicited in a lightweight way before optimization starts: the user specifies a small set of examples on which they expect their judgments to be reliable, which we call \emph{anchors}. These anchors mark regions where comparisons are likely to be stable, without requiring confidence scores after each duel. The question is then how to turn them into a practical heteroscedastic PBO method that accounts for both utility and ease of comparison.

\textbf{Contributions.}
We study PBO under \emph{user-side heteroscedasticity}, where comparison noise depends on how reliable the user is locally in the design space. Our contributions are:
\begin{itemize}[leftmargin=1.2em]
    \item \textbf{Methodological:} We introduce an anchor-based heteroscedastic preference model, where expert-provided anchors define regions of reliable judgment, inducing an input-dependent noise map \emph{via} Kernel Density Estimation (KDE), and we develop risk-aware acquisition functions for this case.

\item \textbf{Theoretical:} We show that a risk-adjusted expected utility of the best option (EUBO) preserves one-step Bayes-optimality up to an additive constant, and establish consistency and concentration for the KDE variance estimator under an idealized i.i.d.\ anchor model.

    \item \textbf{Empirical:} We demonstrate improved risk-adjusted performance on synthetic and real preference benchmarks, and study sensitivity to anchor quality and anchor count.
\end{itemize}

\begin{figure*}[ht!]
    \centering
    \includegraphics[width=1\linewidth]{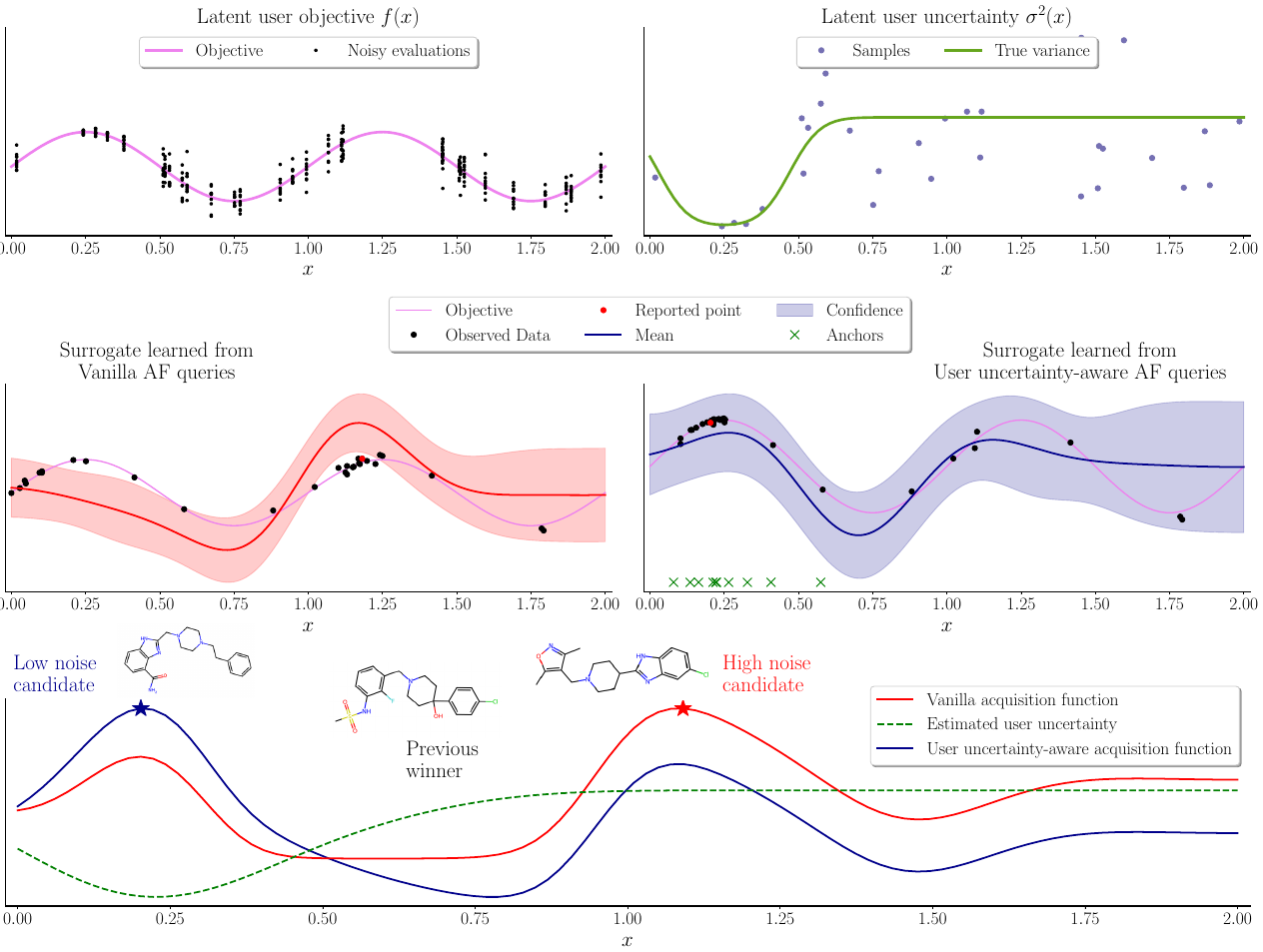}
    % \caption{\textbf{Toy illustration of anchor-based heteroscedastic PBO.} Anchors identify a low-uncertainty region, yielding an input-dependent uncertainty map (green). Acquisition functions that account for this uncertainty query easier comparisons and recover a better-calibrated preferential GP than a risk-neutral baseline.}
    \caption{\textbf{Heteroscedastic preferential Bayesian optimization.}
Top left: latent user utility with heteroscedastic noisy evaluations (Scalar samples shown only to visualize the latent noise and are not observed by PBO.)
Top right: true latent user uncertainty in this toy example. Middle: preferential GP surrogates obtained from queries selected by a vanilla acquisition function (left) and a user-uncertainty-aware acquisition function (right). Our anchor-based model of user uncertainty favors lower-noise yet still promising queries, leading to a better-calibrated surrogate (blue) than the vanilla baseline (red). Importantly, the latent function is inferred from \emph{preferences}, not direct objective evaluations. Bottom: acquisition landscapes. The estimated user uncertainty (green) modifies the acquisition function (blue), shifting its maximizer away from the vanilla one (red) toward a lower-variance candidate, resulting in this illustrative example in two different molecule candidates.}
    % \caption{\textbf{Heteroscedastic Preferential Bayesian Optimization}. Top left: latent user utility with heteroscedastic noisy evaluations. Top right: ground truth latent user uncertainty in this example. Middle: Preferential GP surrogates obtained using queries from vanilla acquisition function (AF) (left) and user-uncertainty-aware AF (right). Our anchor-based model of \emph{user epistemic  uncertainty} leads to queries associated with lower noise and yet similarly high values, resulting in a better-calibrated surrogate (blue) compared to the vanilla GP (red, left). Crucially, the latent function is inferred after observing \emph{human preferences}, not objective evaluations. Bottom: AF landscape. The estimated user epistemic uncertainty (green) informs a user-uncertainty-aware acquisition function (blue), leading to a maximizer that differs from the vanilla AF (red) and accurately selects the low-variance design.}
    \label{fig:gist}
\end{figure*}

\section{Background}\label{sec:background}

The goal of PBO is to identify the optimum
\(\mathbf{x}^\ast = \argmax_{\mathbf{x} \in \mathcal{X}} f(\mathbf{x})\)
from duel feedback \(\x \succ \x'\), signifying a preference of \(\x\) over \(\x'\).
The latent utility function \(f:\mathcal{X}\to\mathbb{R}\) is defined on \(\mathcal{X}\subset\mathbb{R}^d\), and the observed comparisons are denoted by
\(\dataset \triangleq \{\x_k \succ \x_k'\}_{k=1}^m\).

\textbf{Prior.}
We place a zero-mean Gaussian process prior on \(f\),
\begin{equation}
f(\x)\sim\mathcal{GP}(0,l(\x,\x')),
\end{equation}
with kernel \(l:\mathcal{X}\times\mathcal{X}\to\mathbb{R}\).
For a finite set of inputs \(\mathbf{X}\), the corresponding latent values
\(\mathbf{f}\triangleq [f(\x)]_{\x\in\mathbf{X}}\)
follow a multivariate Gaussian distribution with covariance matrix
\(\mathbf{L}\triangleq [l(\x,\x')]_{\x,\x'\in\mathbf{X}}\).

\textbf{Likelihood.}
Given duel observations \(\dataset\), we write
\vspace{-.3cm}
\begin{equation}
p(\dataset\mid \mathbf{f})
=
\prod_{k=1}^m
p(\x_k \succ \x_k' \mid f(\x_k), f(\x_k')).
\end{equation}
\vspace{-.3cm}
The duel likelihood takes one of the following forms:
\begin{align}
p(\x \succ \x' \mid f(\x),f(\x'))
&=
\Phi\!\left(
\frac{f(\x)-f(\x')}
{\sqrt{\sigma_\varepsilon^2(\x)+\sigma_\varepsilon^2(\x')}}
\right),
&& \text{(probit)} \label{eq:probit-like} \\
p(\x \succ \x' \mid f(\x),f(\x'))
&=
\frac{\exp(f(\x)/\lambda(\x))}
{\exp(f(\x)/\lambda(\x))+\exp(f(\x')/\lambda(\x'))},
&& \text{(logistic)} \label{eq:logistic-like}
\end{align}
where \(\sigma_\varepsilon^2(\x)\) and \(\lambda(\x)\) denote local noise levels.
PBO typically assumes the homoscedastic special cases
\(\sigma_\varepsilon^2(\x)=\sigma_{\mathrm{noise}}^2\) or
\(\lambda(\x)=\lambda_{\mathrm{noise}}\); we instead model user uncertainty as input-dependent.

% \textbf{Likelihood.}
% Given duel observations \(\dataset\), we write
% \begin{equation}
% p(\dataset\mid \mathbf{f})
% =
% \prod_{k=1}^m
% p(\x_k \succ \x_k' \mid f(\x_k), f(\x_k')).
% \end{equation}
% Under a probit likelihood, duel outcomes are generated by additive Gaussian noise,
% \begin{equation}
% \x \succ \x'
% \;\Longleftrightarrow\;
% f(\x)+\varepsilon(\x) > f(\x')+\varepsilon(\x'),
% \qquad
% \varepsilon(\x)\sim\mathcal{N}(0,\sigma_\varepsilon^2(\x)).
% \end{equation}
% Under a logistic likelihood,
% \begin{equation}
% p(\x \succ \x' \mid f(\x),f(\x'))
% =
% \frac{\exp(f(\x)/\lambda(\x))}
% {\exp(f(\x)/\lambda(\x))+\exp(f(\x')/\lambda(\x'))},
% \end{equation}
% where \(\lambda(\x)\) is the local noise level.
% Existing PBO methods typically assume the homoscedastic special cases
% \(\sigma_\varepsilon^2(\x)=\sigma_{\mathrm{noise}}^2\) or
% \(\lambda(\x)=\lambda_{\mathrm{noise}}\); we instead model user uncertainty as input-dependent.

\textbf{Posterior.}
Combining the GP prior and the preference likelihood yields the posterior
\[
p(\mathbf{f}\mid\dataset)
=
\frac{p(\dataset\mid\mathbf{f})\,p(\mathbf{f})}
{\int p(\dataset\mid\mathbf{f})\,p(\mathbf{f})\,d\mathbf{f}}.
\]
Because the preference likelihood is non-Gaussian, this posterior is generally intractable and must be approximated.
Standard choices include Laplace, EP, and sampling-based approximations; details for the probit and logistic surrogates used in this work are deferred to Appendices~\ref{app:surrogate}--\ref{app:detail-EP}.

% \textbf{Acquisition function.}
% Given the posterior, PBO selects the next query via an acquisition function.
% A generic single-point BO criterion takes the form
% \begin{equation}\label{eq:af}
%     \x^{\mathrm{next}}
%     =
%     \argmax_{\x\in\mathcal{X}}
%     \alpha(\x)
%     :=
%     \int u(\x,\mathbf{f})\,p(\mathbf{f}\mid\dataset)\,d\mathbf{f},
% \end{equation}
% where \(u\) measures the value of querying \(\x\).
% In PBO, heuristic methods adapt such single-point criteria to construct a duel, while decision-theoretic methods score the whole duel jointly \citep{pmlr-v206-astudillo23a}.
% Our heteroscedastic noise model can be used in both cases.

\textbf{Acquisition function.}
Given the posterior, PBO selects the next duel via an acquisition function.
Some acquisition rules are inherited from single-point BO criteria, of the form
\begin{equation}\label{eq:af}
    \x^{\mathrm{next}}
    =
    \argmax_{\x\in\mathcal{X}}
    \alpha(\x)
    :=
    \int u(\x,\mathbf{f})\,p(\mathbf{f}\mid\dataset)\,d\mathbf{f},
\end{equation}
where \(u\) measures the value of querying \(\x\).
In PBO, such criteria can be used to select a challenger, which is then paired with the previous winner/current incumbent, a common strategy~\citep{takeno2023towards}.
Alternatively, batch decision-theoretic strategies score a full pair
\(\X=[\x_1,\x_2]\) directly,
\[
    \X^{\mathrm{next}}
    =
    \argmax_{\X\in\mathcal{X}^2}
    \alpha(\X),
\]
%as in batch EI adaptations~\citep{siivola2021preferential} or
as in EUBO~\citep{pmlr-v206-astudillo23a}.
In this case, the two duel elements are optimized jointly, rather than obtained by enumerating all possible pairs.
Our heteroscedastic noise model can be used in both single-point and pair-valued acquisition rules.

\section{Preferential BO with heteroscedastic noise}\label{sec:PBO-heteroscedastic}

We model user uncertainty in the heteroscedastic PBO setting of \Cref{sec:background}. \Cref{sec:hnd} introduces the input-dependent noise model, \Cref{subsec:simplemodel} shows how anchors yield a density estimate, and \Cref{subsec:risk-averse-af} incorporates the resulting uncertainty map into risk-aware acquisition functions.

\vspace{-.5cm}

\subsection{Heteroscedastic noise distribution}\label{sec:hnd}

For probit surrogates, we replace the constant noise variance by
\begin{equation}\label{eq:general-noise}
    \varepsilon(\x) \sim \mathcal{N}(0, \sigma_\varepsilon^2(\x) = a \exp(- q(\x))),
\end{equation}
and for logistic surrogates we analogously use
\begin{equation}\label{eq:general-logistic-noise}
 \lambda(\x) = a \exp(- q(\x)),
\end{equation}
where \(q(\x)\geq 0\) is a reliability score and \(a>0\) sets the overall noise scale. Larger values of \(q(\x)\) correspond to lower comparison noise. If \(q(\x)\) is uniform, the model reduces to the usual homoscedastic likelihood. In our anchor-based model below, \(q\) controls the spatial variation of user uncertainty, while \(a\) calibrates its global magnitude.
%
% with a scaling factor $a>0$ and a non-negative function $q(\x)$ that we interpret as a reliability score. If $q(\x)$ is uniform, the model reduces to the usual homoscedastic likelihood.
% The function \(q(\x)\) controls the spatial variation of the uncertainty map, while \(a\) sets its global scale. Thus, anchors determine where the user is relatively more or less reliable, whereas \(a\) calibrates the overall magnitude of the comparison noise. In all experiments we fix \(a=1\) and select the KDE bandwidth \(h\) by leave-one-out cross-validation.

\subsection{\emph{Anchors}-based input-dependent noise}\label{subsec:simplemodel}

We introduce \emph{anchors}, a set of reliability markers $\mathbf{X}_0 = \{\x_i\}_{i=1}^n$ provided by the expert, around which judgments are expected to be reliable. Given $\X_0$, we estimate $q(\x)$ with a KDE,
\begin{equation}
    \hat{q}(\mathbf{x} \vert h, \X_0) = \frac{1}{n} \sum_{i=1}^n \frac{1}{h^d} \, k \left( \frac{\Vert \mathbf{x} - \mathbf{x}_i \Vert_2}{h} \right),
\label{eq:kde}
\end{equation}
where $k$ is a kernel and $h$ a bandwidth. A high anchor density implies low comparison noise through \Cref{eq:general-noise,eq:general-logistic-noise}. The estimator depends only on anchor locations and can therefore be constructed before the PBO loop.

In standard heteroscedastic BO, a separate GP over the observation noise is natural because repeated scalar observations help disentangle signal and noise. In PBO, binary comparisons are less informative, so utility variation and user uncertainty are more easily confounded. Anchors instead provide external information about \emph{where} comparisons are reliable, and KDE turns this side information into an uncertainty map. This also matches a case-based view of expert judgment, where unfamiliar inputs are assessed by similarity to known cases. We treat anchors as exchangeable user-supplied prototypes; the stronger i.i.d.\ assumption is only needed for the estimator analysis in \Cref{sec:theory}.

\subsection{Risk-averse acquisition functions}\label{subsec:risk-averse-af}

We interpret heteroscedastic user uncertainty as a source of \emph{risk}: when the local noise level is high, a queried duel is less reliable and thus less useful for optimization. Following prior work on risk-sensitive decision making and risk-averse BO \citep{kahneman2013prospect,hull2012risk,golub2000risk,makarova2021risk,kuindersma2013variable}, we favor acquisition functions that trade off high latent utility against low aleatoric uncertainty. In our setting, this uncertainty is quantified by the input-dependent noise variance estimated by the heteroscedastic preference model.

\textbf{Heuristic acquisition functions.}
For heuristic PBO rules, we penalize candidates with large estimated noise variance \(\sigma_\varepsilon^2(\x)\).
The first rule is the \emph{aleatoric noise-penalized expected improvement} (ANPEI), which adapts EI to the preferential setting by replacing the incumbent value with the best posterior mean over previously queried points:
\begin{equation}
\alpha_{\mathrm{ANPEI}}(\x)
=
\mathbb{E}\!\left[
\bigl(
\mu_{\ast\mid\dataset}(\x)-\mu_{\ast\mid\dataset}(\x^\ast)
\bigr)_+
\right]
-
\gamma \sigma_\varepsilon(\x),
\label{eq:anpei}
\end{equation}
where \(\gamma>0\) controls the strength of the penalty.
Here \(\mu_{\ast\mid\dataset}\) and \(\sigma_{\ast\mid\dataset}^2\) denote the posterior mean and variance of the preferential GP surrogate, and \(\x^\ast\) is the previously queried point with largest posterior mean.
The second rule is the \emph{risk-averse upper confidence bound} (RAHBO). For \(\eta,\gamma>0\),
\begin{equation}
    \alpha_{\mathrm{RAHBO}}(\x)
    =
    \mu_{\ast \vert \dataset}(\x)
    +
    \eta \sigma_{\ast \vert \dataset}(\x)
    -
    \gamma \sigma_\varepsilon^2(\x),
    \label{eq:rucb}
\end{equation}
%gamma>0\). RAHBO retains the optimism from UCB while explicitly discouraging noisy queries.
%
\textbf{Decision-theoretic acquisition.}
For decision-theoretic PBO, we build on EUBO \citep{pmlr-v206-astudillo23a}, which scores a duel through the expected utility of its best element. To make this criterion risk-averse, we replace utility by the mean-variance style quantity
\(\mathrm{MV}(\x)\triangleq f(\x)-\alpha\lambda(\x)\),
where \(\lambda(\x)\) is the local noise level under the logistic preference model. This yields
\begin{equation}\label{eq:raeubo}
\alpha_\mathrm{RAEUBO}(\X)
=
\mathbb{E}_m[\max\{ \mathrm{MV}(\x_1), \mathrm{MV}(\x_2) \}],
\end{equation}
with \(\X=[\x_1,\x_2]\) and \(\mathbb{E}_m\) the conditional expectation given the current dataset.
RAEUBO thus scores the \emph{whole duel} jointly, favoring pairs with high latent utility and low user uncertainty.

Because the penalty terms in \eqref{eq:anpei}--\eqref{eq:raeubo} are additive, the resulting policies naturally prefer regions where the user is expected to be more consistent, which in practice often occurs near the anchors.
This is related to \(\pi\)BO \citep{hvarfner2022pibo}, where prior information is injected directly into the acquisition function using a multiplicative prior term,
\(\alpha_{\pi\mathrm{BO}}(\x)\triangleq \alpha(\x)\pi(\x)\),
whereas our approach uses an additive penalty induced by the estimated heteroscedastic noise.

\section{Theoretical Analysis}\label{sec:theory}

Our theoretical results support two components of the method. First, we show that RAEUBO retains a one-step decision-theoretic interpretation after replacing utility by its risk-adjusted counterpart. Second, for the anchor-based uncertainty estimator, we derive consistency and concentration results as the number of anchors grows, under an idealized i.i.d.\ anchor model.

\subsection{Risk-averse one-step Bayes optimal policy of RAEUBO}

Given $\X = [\x_1, \x_2]$, define the one-step risk-adjusted value
\begin{equation}
\hat{V}_n^\lambda(\X) = \mathbb{E}_n\left[\max_{\x \in \X} \mathbb{E}_{n}[\mathrm{MV}(\x) \vert \X, r(\X)] \right].
\end{equation}
Here \(\mathbb{E}_n\) denotes expectation conditional on the current dataset \(\mathcal{D}_n\), and \(r(\X)\in\{1,2\}\) denotes the index of the winner of the queried pair \(\X=(\x_1,\x_2)\).
This is the heteroscedastic counterpart of the $q$EUBO value analyzed by \citet{pmlr-v206-astudillo23a}.

\begin{restatable}{proposition}{RA-one-step-bayes}\label{prop:one-step-bayes}
Suppose human feedback follows the logistic likelihood with a heteroscedastic noise model $\lambda(\x), \forall \x \in \mathcal{X}$. If $ \X_\ast \in \argmax_{\X \in \mathcal{X}^2} \alpha_\mathrm{RAEUBO}(\X)$, then there exists a constant $\hat{\lambda} > 0$ s.t.
\begin{equation}
\hat{V}_n^\lambda(\X_\ast) \geq \max_{\X \in \mathcal{X}^2} \hat{V}_n^0(\X) - \hat{\lambda} C
\end{equation}
for $C = L_W((q - 1) / e)$ and $L_W$ the Lambert $W$ function.
\end{restatable}
Thus, maximizing RAEUBO recovers a one-step policy that is optimal up to an additive constant. The proof is deferred to Appendix~\ref{app:proof-one-step-bayes}.

\subsection{Consistency analysis of the KDE-based model of user epistemic uncertainty}\label{subsec:consistency-analysis}

For the estimator analysis, we study the discrepancy between the estimated local noise variance and the true variance through its mean squared error.
To invoke standard KDE consistency results, we strengthen the practical exchangeability view of \Cref{subsec:simplemodel} and assume that the anchors are i.i.d.
We also impose the usual regularity conditions on the underlying density $q$ and variance functions:

\begin{assumption}\label{assumption:risk-analysis}
$q$ belongs to a class of densities $\mathcal{P}(\beta, L)$ defined by
\begin{align*}
    \mathcal{P}(\beta, L) \triangleq \left\{ q : q \geq 0, \int q(\x) \mathrm{d}\x = 1,\; q \in \Sigma(\beta, L) \text{ on } \mathbb{R}^d \right\},
\end{align*}
with $\Sigma(\beta, L)$ the Hölder class.
\end{assumption}

Under this smoothness assumption, the anchor-based estimator converges at the usual KDE rate:

\begin{restatable}{proposition}{riskanalysisproposition}\label{prop:mse}
    Fix $\alpha > 0$ and take $h = \alpha n^{-1/(2\beta + d)}$. Then, for any input $\x$ and any number of anchors $n \geq 1$, the estimated variance satisfies
    \begin{equation}
        \underset{q \in \mathcal{P}(\beta, L)}{\sup} \mathbb{E}_{\mathbf{X_0}}\!\left[(\hat{\sigma}_\varepsilon^2(\x) - \sigma_\varepsilon^2(\x))^2\right] \leq a^2 c_3 n^{- \frac{2 \beta}{2 \beta + d}},
    \end{equation}
    where $c_3 > 0$ is a constant depending on $\beta$, $\alpha$, $a$, and the kernel bandwidth.
\end{restatable}

The rate degrades with dimension, which already suggests that anchor-based KDE is primarily useful in low or moderate dimensions.
Because the i.i.d.\ assumption is stronger than our practical anchor-elicitation model, we complement the theory with robustness experiments under few-anchor and non-i.i.d.\ settings (Section~\ref{sec:experiments} and Figure~\ref{fig:optima-results}).
Of note, this convergence is with respect to the number of anchors, not the number of PBO iterations.
Since the KDE map is fixed during a given optimization run, 
the result quantifies how many anchors are needed to accurately recover the underlying user-uncertainty.

Finally, we complement the MSE analysis with a concentration result that controls the deviation of \(\hat{\sigma}_\varepsilon^2(\x)\) from the true variance \(\sigma_\varepsilon^2(\x)\). Under the same assumptions as above, we obtain:

% \begin{restatable}{proposition}{concentrationanalysisproposition}\label{prop:concentration-inequality}
% For any $\delta > 0$ and $a \geq 1$, there exist constants $c_1$ and $c_2$ such that
% %
% \begin{equation}\label{eq:variance-bernstein-inequality}
% \underset{q \in \mathcal{P}(\beta, L)}{\sup}\; \vert \sigma_\varepsilon^2(\x) - \hat{\sigma}^2_\varepsilon(\x) \vert > a \left( \sqrt{\frac{4c_1}{nh^d} \log \frac{2}{\delta}} + c_2 h^{\beta} \right)
% \end{equation}
% %
% with probability less than $\delta$ for any $\x$. Furthermore, under an additional VC-type assumption on the kernel class, a bound that holds uniformly over $\x \in \mathcal{X}$ also holds:
% %
% \begin{equation}\label{eq:variance-gine-inequality}
% \underset{\x \in \mathcal{X}}{\sup}\; \vert \sigma_\varepsilon^2(\x) - \hat{\sigma}^2_\varepsilon(\x) \vert > a \left( \sqrt{ \frac{1}{c_4 \, n \, h_n^d } \, \log \frac{c_3}{\delta}} + c_2 h^{\beta} \right),
% \end{equation}
% %
% for constants $c_3,c_4$ and bandwidth sequence $h_n$.
% \end{restatable}

\begin{restatable}{proposition}{concentrationanalysisproposition}\label{prop:concentration-inequality}
For any \(\delta>0\) and \(a\geq 1\), there exist constants \(c_1,c_2>0\) such that, for any fixed \(\x\),
\begin{equation}
\sup_{q\in\mathcal{P}(\beta,L)}
\mathbb{P}\!\left(
\left|\sigma_\varepsilon^2(\x)-\hat{\sigma}_\varepsilon^2(\x)\right|
\le
a\left(
\sqrt{\frac{4c_1}{nh^d}\log\frac{2}{\delta}}
+
c_2h^\beta
\right)
\right)
\leq 1-\delta .
\label{eq:variance-bernstein-inequality}
\end{equation}
Furthermore, under an additional VC-type assumption on the kernel class, there exist constants \(c_3,c_4>0\) and a bandwidth sequence \(h_n\) such that
\begin{equation}
\sup_{q\in\mathcal{P}(\beta,L)}
\mathbb{P}\!\left(
\sup_{\x\in\mathcal{X}}
\left|\sigma_\varepsilon^2(\x)-\hat{\sigma}_\varepsilon^2(\x)\right|
\le
a\left(
\sqrt{\frac{1}{c_4nh_n^d}\log\frac{c_3}{\delta}}
+
c_2h_n^\beta
\right)
\right)
\leq 1-\delta .
\label{eq:variance-gine-inequality}
\end{equation}
\end{restatable}

The first bound gives pointwise high-probability control, the second strengthens this to uniform control over \(\x\) under additional VC-type and bandwidth assumptions. Proofs are given in Appendix~\ref{app:proof-of-concentration}.

\section{Related work}\label{sec:related-works}

\textbf{Bayesian optimization with heteroscedastic noise.}
In most real-world settings, defining the likelihood based on a fixed variance noise leads to misspecified posteriors, which can harm the optimization by suggesting designs incorrectly believed to yield high function values. Several attempts have been made to tackle this issue at acquisition level~\citep{makarova2021risk}, surrogate level~\citep{hebo,griffiths2021achieving} or by adapting distribution-free uncertainty quantification techniques like conformal inference to BO~\citep{pmlr-v206-stanton23a}. Lastly, recent work studied the heteroscedastic tradeoff between evaluating fewer conditions with more replicates versus more conditions with fewer replicates \citep{dai2024batch}.

\textbf{Preferential Bayesian optimization.}
As in standard BO, PBO is largely determined by the surrogate and acquisition function.
Since its posterior is generally intractable, prior work has relied on Laplace approximation, skew-GP formulations, EP, and MCMC~\citep{brochu2010tutorial,benavolipbo,takeno2023towards}.
Neural networks have also been used instead of GPs~\citep{daolangpaper}.
On the acquisition side, some work adapted classical strategies to PBO by selecting one of the duel elements as the winner of the previous duel and optimizing for the second element using Expected Improvement or Thompson Sampling~\citep{siivola2021preferential}. Next, acquisition functions leveraging the preferential nature of the queries were also proposed~\citep{gonzalez2017preferential,pmlr-v206-astudillo23a}.
Recent work proposed theoretically grounded PBO algorithms, with regret bounds~\citep{principledpbo, pasztor2024bandits, kayal}.
More recently, amortized PBO has been proposed to meta-learn both the surrogate and acquisition function, offering a computationally cheaper alternative to GP-based PBO~\citep{zhang2025pabbo,zhang2026taskagnostic}.
Lastly, preferential relations have been leveraged to enhance vanilla BO: ~\cite{hvarfner2024a} investigate a user-defined prior integrated to
the GP surrogate, and~\citep{loopinghuman} allow the prior to be iteratively updated.

\section{Experiments}\label{sec:experiments}

\begin{figure*}[h!]
    \centering
    \includegraphics[width=.9\linewidth]{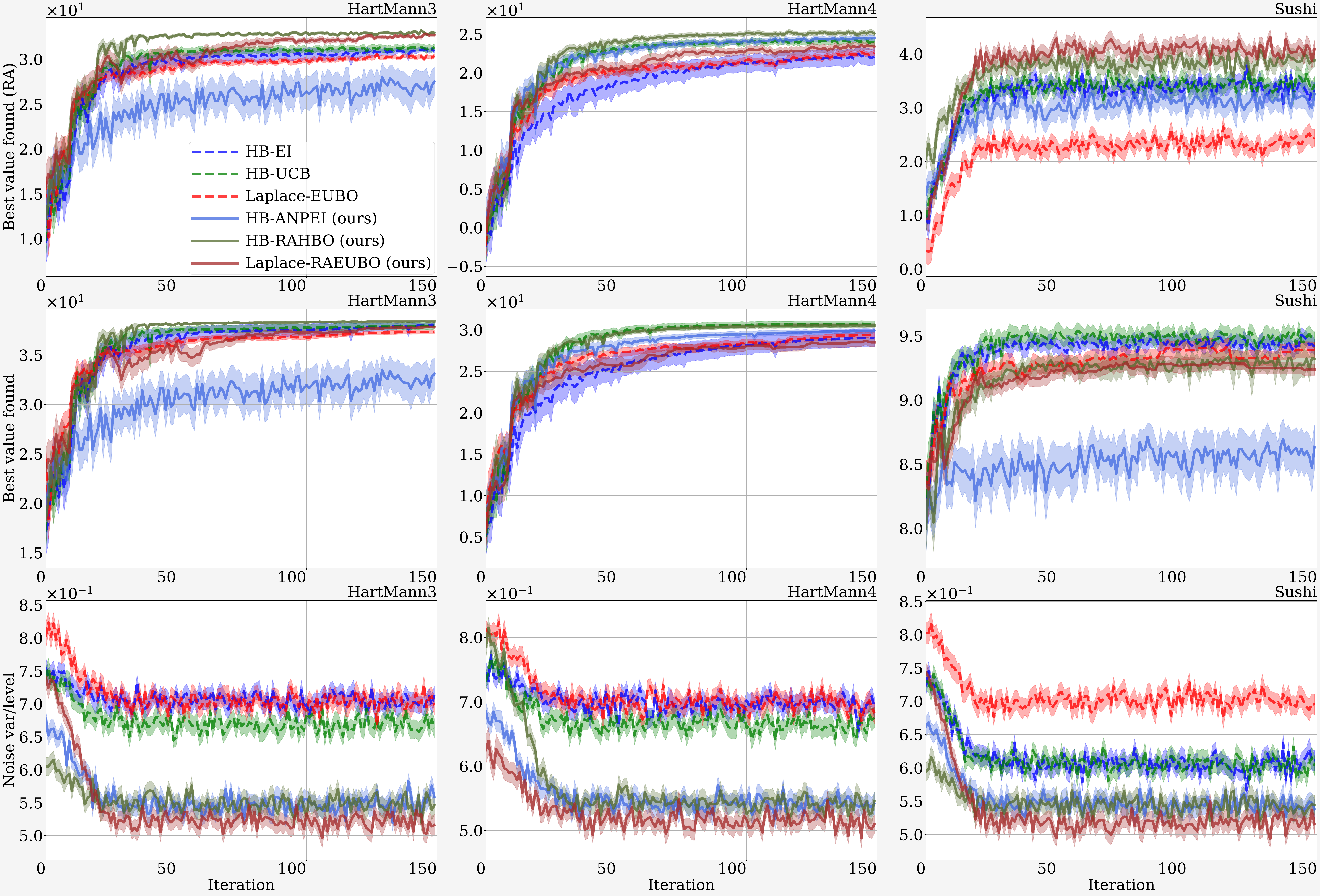}
    \caption{Main results. Top: risk-adjusted best value. Middle: conventional best value. Bottom: noise level of queried pairs. Mean $\pm 0.2$ std over 30 runs.  \textbf{Risk-aware strategies improve the primary risk-adjusted objective and usually query less noisy pairs.}}%Ù  on the conventional best-value metric they remain competitive.}}
    \label{fig:synthetic-results}
\end{figure*}

We evaluate whether modeling user uncertainty changes which duels should be queried across synthetic and real preference benchmarks. Our study covers classical BO test functions (Hartmann3D, Hartmann4D, and Ackley6D) as well as real-world examples (Sushi and Candy,~\citealt{siivola2021preferential}), together with ablations on anchor quality, anchor count and non-i.i.d.\ anchor sampling. % and alternative inference schemes.

\begin{figure}[h!]
    \centering
    \includegraphics[width=.8\linewidth]{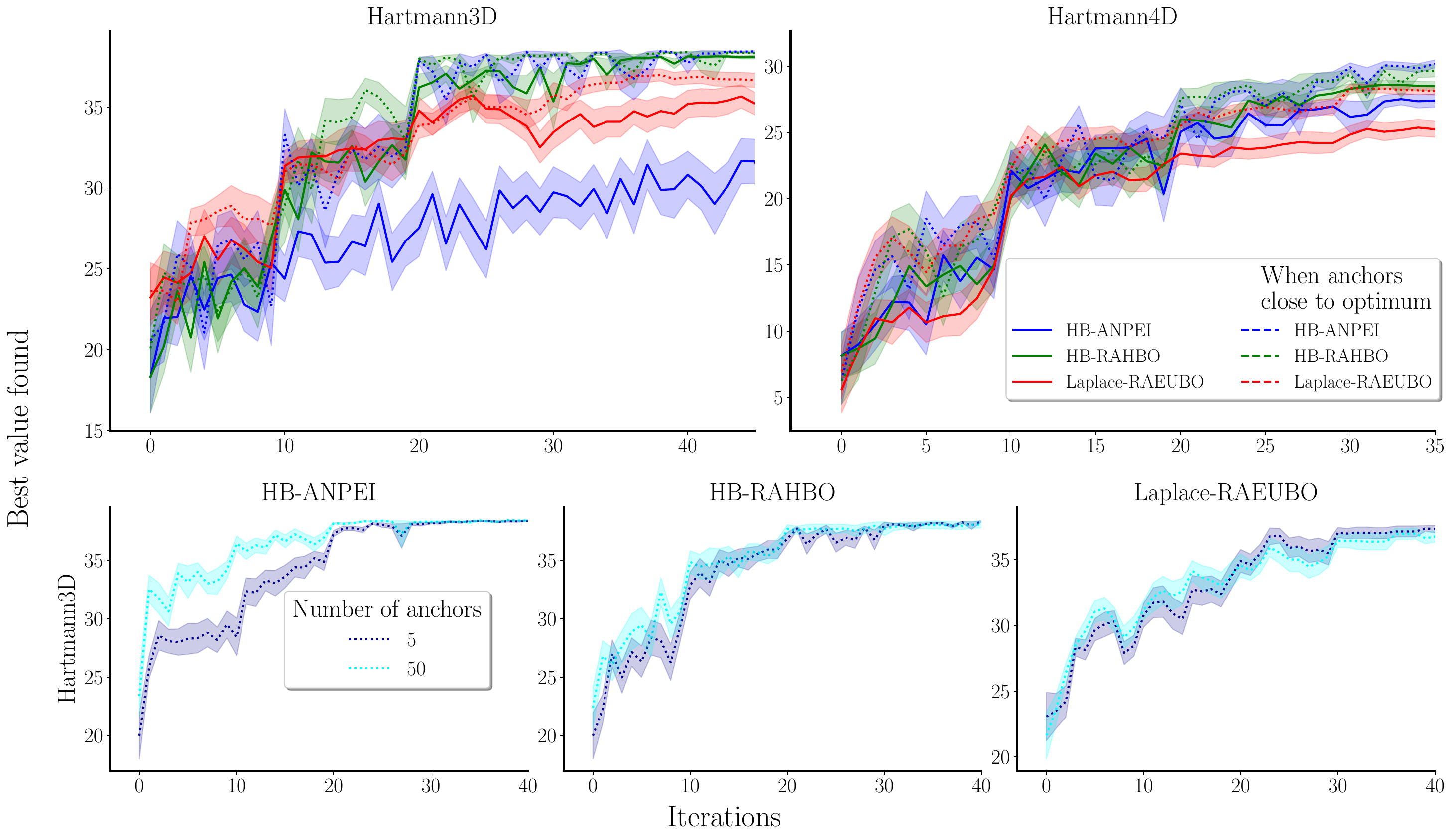}
    \includegraphics[width=.8\linewidth]{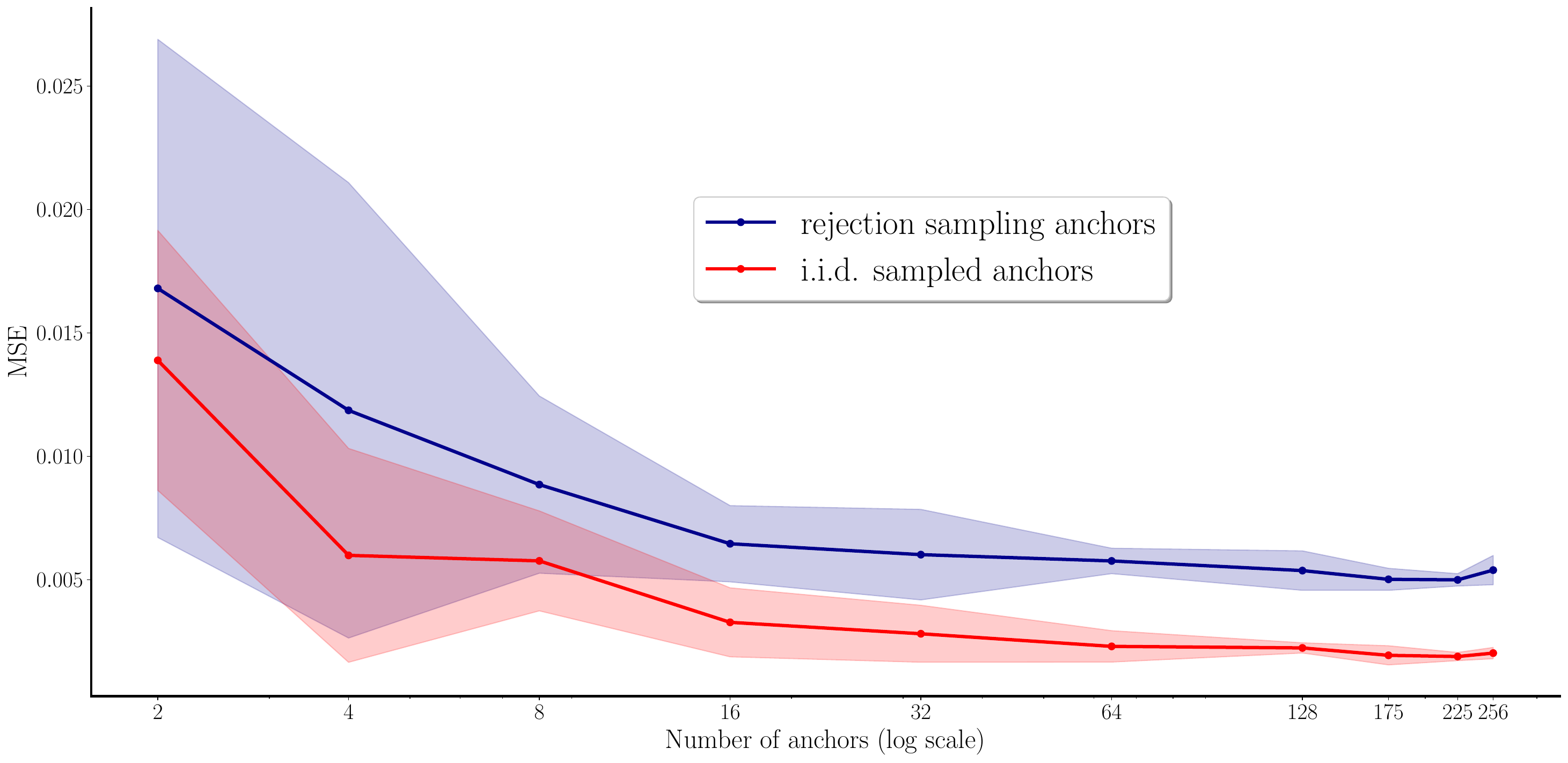}
    
\caption{Anchor sensitivity.
\textbf{Top:} effect of anchor quality on optimization for Hartmann3D and Hartmann4D. \textbf{Placing anchors closer to the optimum accelerates convergence}, as the low-uncertainty region better overlaps with the high-utility region.
\textbf{Middle:} effect of anchor count on Hartmann3D for HB-ANPEI, HB-RAHBO, and Laplace-RAEUBO; \textbf{using more anchors improves performance}.
\textbf{Bottom:} consistency of the KDE-based noise estimator on a 1D problem as a function of the number of anchors, comparing i.i.d.\ and rejection-sampling anchor processes. Although our theory assumes i.i.d.\ anchors, \textbf{the estimator remains effective under rejection sampling}.
Mean $\pm 0.2$ std over 30 random seeds.}
    \label{fig:optima-results}
\end{figure}

\textbf{Setup.}
We compare the risk-neutral baselines EI, UCB, and EUBO with their noise-aware counterparts ANPEI, RAHBO, and RAEUBO. 
In all experiments, we fix \(a=1\) in the heteroscedastic noise model (Equations~\ref{eq:general-noise}--\ref{eq:general-logistic-noise}) and select the KDE bandwidth \(h\) by leave-one-out cross-validation (Equation~\ref{eq:kde}; details in \Cref{app:hyperparameter-optimization}).
The acquisition penalty parameters are fixed across experiments, using \(\gamma=\alpha=10\) and \(\eta=2\) (Equations~\ref{eq:anpei}--\ref{eq:raeubo}).
Probit-based methods use Hallucination Believer inference~\citep{takeno2023towards}, while logistic ones use Laplace approximation~\citep{pmlr-v206-astudillo23a}; the heteroscedastic likelihood can be combined with different posterior approximations, with HB, Laplace, and EP formulations given in the appendix.
Surrogate hyperparameters are fit by marginal likelihood maximization.

% \paragraph{Setup.}
% We compare the risk-neutral baselines EI, UCB, and EUBO with their noise-aware counterparts ANPEI, RAHBO, and RAEUBO. 
% In all experiments, we fix \(a=1\) (Equation~\ref{eq:general-noise}) and select the KDE bandwidth \(h\) by leave-one-out cross-validation (Equation~\ref{eq:kde}).
% The acquisition penalty parameters are fixed across experiments, using \(\gamma=\alpha=10\) and \(\eta=2\) (Equations~\ref{eq:anpei} and~\ref{eq:rucb}).
% Probit-based methods use Hallucination Believer inference \citep{takeno2023towards}; logistic ones use Laplace approximation \citep{pmlr-v206-astudillo23a}. Surrogate hyperparameters are fit by marginal likelihood maximization, and the bandwidth selection for our KDE-based uncertainty model is detailed in \Cref{app:hyperparameter-optimization}. 

\textbf{Metrics.} Our primary metric is the risk-adjusted best value
\(f(\x_t^\ast) - \rho \, n(\x_t^\ast)\) with \(\rho=10\), where
\(\x_t^\ast \triangleq \argmax_{\x \in \mathcal{X}} \mu_{\ast \vert \dataset}(\x) - \rho n(\x)\).
Here, \(n(\x)=\hat{\sigma}^2_\varepsilon(\x)\) for probit surrogates and \(n(\x)=\lambda(\x)\) for logistic ones.
We also report the conventional best-value metric \(f(\x_t^\ast)\), together with the noise level of queried pairs.
The risk-adjusted metric is a complementary diagnostic, not a replacement for the conventional best-value objective. It measures whether a method finds candidates that are both promising and expected to be reliably compared by the user, which matters when similar-utility candidates differ in comparison difficulty and cognitive load.
%The appendix reports ranking variability and computation times.

\textbf{Synthetic functions.}
To stress-test the method, synthetic anchors are placed away from the optimum, so the low-uncertainty region does not coincide with the most valuable part of the objective landscape. Even in this case, the risk-aware AFs improve the risk-adjusted objective on both Hartmann problems (Figure~\ref{fig:synthetic-results}, top row). On the conventional best-value metric they remain competitive, while querying systematically lower-noise pairs (bottom row). RAHBO and RAEUBO provide the most stable trade-off, whereas ANPEI is more sensitive to mismatch between low-noise and high-utility regions. On the higher-dimensional Ackley 6D task (Figure~\ref{fig:high-dim-results}), only Laplace-RAEUBO remains clearly competitive, consistent with the known deterioration of KDE in higher dimensions.

\textbf{Real-world examples: learning human preferences on Sushi and Candy.}
Following \citet{siivola2021preferential}, we turn the 100-item Sushi dataset into a continuous 4D benchmark with a \(3\)-nearest-neighbors regressor. This benchmark construction is inherited from prior PBO work; our contribution is the uncertainty model on top of it. Using 13 representative sushi types as anchors, RAEUBO achieves the best risk-adjusted performance and RAHBO is second (Figure~\ref{fig:synthetic-results}, right column). Both remain competitive on the conventional best-value metric while selecting less noisy pairs than their risk-neutral counterparts.
We additionally evaluate on a real-world Candy dataset with 85 items, each described by sugar percentage and price percentage, using 12 chocolate-based candies as anchors. On this task, our risk-aware methods again improve over the risk-neutral baselines on both the risk-adjusted and conventional best-value metrics, with the exception of RAEUBO (Figure~\ref{fig:real-world-results}). %This second real-world result suggests that the benefits of modeling user-side heteroscedasticity extend beyond the Sushi benchmark.

\textbf{Ablation studies on anchors.}
Figure~\ref{fig:optima-results} highlights three practical aspects of the anchor model. First, anchors closer to the optimum accelerate convergence, since the low-uncertainty region better overlaps with the high-utility region (top row). Second, even when anchors are suboptimal, increasing their number improves performance, especially in early iterations (middle row). Third, the bottom panel shows that although our KDE estimator analysis assumes i.i.d.\ anchors, the estimator remains effective under a non-i.i.d.\ rejection-sampling anchor process, with error decreasing in a similar way as the number of anchors grows.

\section{Discussion and limitations}\label{sec:conclusions}

Anchor-based heteroscedastic noise provides a simple way to inject human-side uncertainty into preferential Bayesian optimization. Empirically, the main effect is not to increase raw utility at any cost, but to redirect the search toward informative and reliable comparisons. This improves the risk-adjusted objective while keeping the conventional best-value metric competitive.

\textbf{Limitations.} Our analysis provides principled support for the proposed approach under a simplified theoretical setting. Extending these guarantees to the full heteroscedastic PBO pipeline under the weaker practical assumption of exchangeable anchors would require additional assumptions and is left for future work. Other natural directions include automatic anchor proposal or validation, and more flexible local-bandwidth noise models.%, and broader comparisons with recent principled risk-neutral PBO methods.
% Let us also mention the multiple expert setting where both the latent utility and the reliability map are user-specific. Multi-user expertise aggregation is an important extension, especially when user feedback is used only as a proxy for an external black-box objective.

\section{Acknowledgments}
This research was supported by the Research Council of Finland (flagship
programme: Finnish Center for Artificial Intelligence, FCAI grants 358958, 345604, and 341763), and
the UKRI Turing AI World-Leading Researcher Fellowship, EP/W002973/1. We also acknowledge
the computational resources provided by the Aalto Science-IT Project.
\bibliography{references}

\clearpage

\appendix

\setcounter{section}{0}
\setcounter{figure}{0}
\setcounter{equation}{0}
\setcounter{table}{0}
\renewcommand\thesection{\Alph{section}}
\renewcommand{\thetable}{S\arabic{table}}
\renewcommand{\thefigure}{S\arabic{figure}}
\renewcommand{\theequation}{S\arabic{equation}}

\textbf{Outline.}

The appendix is organized as follows:
\begin{itemize}[leftmargin=1.2em]
    \item \Cref{app:surrogate} details the surrogate model, including the probit and logistic likelihoods, hyperparameter optimization, and the complexity of the KDE-based noise estimator.
    
    \item \Cref{app:HB,app:detail-LA,app:detail-EP} describe the approximate inference schemes used in this work: Hallucination Believer, Laplace approximation, and expectation propagation.
    
    \item \Cref{app:proofs} contains the proofs of the main theoretical results, including the one-step guarantee for the risk-adjusted acquisition rule and the consistency and concentration analysis of the KDE-based estimator.
    
    \item \Cref{app:addexp} reports additional experimental results, including estimator-consistency studies, ranking plots, computation times, the five-anchor regime, the Candy benchmark, and the Ackley6D experiment.
    
    \item \Cref{app:id} provides implementation details and a summary of the experimental setup.
\end{itemize}

\section{Details of surrogate model}\label{app:surrogate}

In this section, we provide the details of the GP surrogate of PBO. Specifically, we discuss the likelihood function, the hyperparameter optimization procedure, and the time complexity of our proposed noise model.

\subsection{Likelihood}\label{app:likelihoods} 

We consider two likelihoods: probit and logistic likelihoods.

\subsubsection{Probit likelihood}

%We first derive the analytic form of the \textbf{probit likelihood}.
As detailed in \citet{chu2005preference}, probit likelihood implies that the duel outcome is determined by:
\begin{equation}
    \x \succ \x' \iff  f(\x) + \varepsilon(\x) > f(\x') + \varepsilon(\x'), 
\end{equation}
where $\varepsilon(\x)$ denote zero-mean additive noise drawn from a normal distribution $\normal(0, \sigma_\varepsilon^2(\x))$, for all input $\x$. Given that $\x_k \succ \x'_k$, we model the probit likelihood $\Phi(\mathbf{z}_k)$ as follows: 
\begin{align}\label{eq:simple-likelihood}
     p(\mathbf{x}_k \succ \mathbf{x}^\prime_k \vert f(\mathbf{x}_k), f(\mathbf{{x}}^\prime_k)) &= p(\mathbf{x}_k \succ \mathbf{x}^\prime_k \vert f(\mathbf{x}_k), f(\mathbf{{x}}^\prime_k),\varepsilon(\x_k),\varepsilon(\x^\prime_k)) \nonumber \\
     &= p(f(\x_k) + \varepsilon(\x_k) - f(\x^\prime_k) - \varepsilon(\x^\prime_k) \geq 0) \nonumber \\
     &= p(\varepsilon(\x^\prime_k) - \varepsilon(\x_k) \leq f(\x_k) - f(\x^\prime_k)) \nonumber \\
     &= p \left( \varepsilon - \varepsilon \leq \frac{f(\x_k) - f(\x^\prime_k)}{\sqrt{a \exp(-  \, \hat{q}(\mathbf{x}_k \vert h, \X_0)) + a \exp(-  \, \hat{q}(\mathbf{x}^\prime_k \vert h, \X_0))}} \right) \nonumber \\
     &=\Phi \left( \mathbf{z}_k \triangleq \frac{f(\mathbf{x}_k) - f(\mathbf{x}^\prime_k)}{\sqrt{a \exp(-  \, \hat{q}(\mathbf{x}_k \vert h, \X_0)) + a \exp(-  \, \hat{q}(\mathbf{x}^\prime_k \vert h, \X_0))}} \right).
\end{align}
Here, $a \exp(- \hat{q}(\mathbf{x}_k \vert h, \X_0))$ and $a \exp(- \hat{q}(\mathbf{x}^\prime_k \vert h, \X_0))$ denote the estimate variance of noise $\varepsilon(\x_k)$ and $\varepsilon(\x^\prime_k)$, given a kernel bandwidth $h$. $\Phi$ is the standard Normal distribution c.d.f.\@

Denoting $v_k \triangleq f(\x_k') + \varepsilon(\x_k') - f(\x_k) - \varepsilon(\x_k)$, we have $\mathbf{v}_m < 0 \triangleq \{ v_k < 0 \}_{k=1}^m$. Recall the notation from the main text $\mathbf{X} \triangleq [\x_{1},\dots,\x_{m},\x^\prime_{1},\dots, \x^\prime_{m}]^\top \in \mathbb{R}^{2m\times d}$ for the concatenation of winners and losers from duels $\mathbf{v}_m$, and $\mathbf{f} \triangleq [f(\x)]_{\x \in \mathbf{X}}$. The likelihood function $p(\dataset \vert \ff)$ is defined as follows:
\begin{align}
\label{eq:likelihood}
p(\dataset \vert \ff) &= p(\mathbf{v}_m < 0 \vert \mathbf{f}) \nonumber \\
&= \prod_{k = 1}^m \Phi \left ( \z_k  \right).
\end{align} 

\subsubsection{Logistic likelihood}

% We next discuss the \textbf{logistic likelihood}.
%Let $r(\x_k, \x'_k)$ be human duel feedback. $r(\x_k, \x'_k) = 1$ implies $\x_k \succ \x'_k$ and $r(\x_k, \x'_k) = 2$ otherwise.
Similarly to the probit likelihood, we make the assumption that the duel outcome may not always be consistent with the latent function $f$. This inconsistency is now characterized by the input-dependent noise level $\lambda(\x)$, giving rise to a logistic likelihood:
\begin{align}
    p(\mathbf{x}_k \succ \mathbf{x}^\prime_k \vert f(\mathbf{x}_k), f(\mathbf{{x}}^\prime_k)) = \frac{\exp(f(\x_k) / \lambda(\x_k))}{\exp(f(\x_k) / \lambda(\x_k)) + \exp(f(\x'_k) / \lambda(\x'_k))}.
\end{align}
Using the same notations as in the probit likelihood setting, we define the likelihood function $p(\dataset \vert \ff)$ as follows:
\begin{equation}
 p(\dataset \vert \ff) = \prod_{k = 1}^m   \frac{\exp(f(\x_k) / \lambda(\x_k))}{\exp(f(\x_k) / \lambda(\x_k)) + \exp(f(\x'_k) / \lambda(\x'_k))}. 
\end{equation}
\subsection{Hyperparameter optimization}\label{app:hyperparameter-optimization}
We minimize the negative log marginal likelihood $p(\mathcal{D} \vert \theta)$ approximated with Laplace approximation for GP hyperparameter optimization. Following \citet{chu2005preference}, the approximation requires obtaining $\mathbf{f}_\mathrm{MAP} = \arg \min_{\mathbf{f}} - \log p(\mathbf{f} \vert \mathbf{v}_m) \approx \arg \min_\mathbf{f} S(\mathbf{f})$ where we define $S(\mathbf{f})$ as 
\begin{align}\label{eq:simple-MAP}
    S(\mathbf{f}) = - \sum_{k = 1}^m \log p(\x_k \succ \x^\prime_k \vert f(\x_k), f(\x^\prime_k)) + \frac{1}{2} \mathbf{f}^\top \mathbf{L}^{-1} \mathbf{f}
\end{align}
For the \textbf{probit likelihood} function, we find the first and second derivatives are given by
\begin{align}
    & \frac{\partial}{\partial \ff} \, S(\mathbf{f}) = 
    \begin{bmatrix}
         -\frac{1}{ \sqrt{ a \exp(- \hat{q}(\mathbf{x}_{1:m} \vert h, \X_0)) + a \exp(- \hat{q}(\mathbf{x}^\prime_{1:m} \vert h, \X_0)) } } \, \cdot \, \frac{\phi(\mathbf{z}_{1:m})}{\Phi(\mathbf{z}_{1:m})} \\
         \frac{1}{ \sqrt{ a \exp(- \hat{q}(\mathbf{x}_{1:m} \vert h, \X_0)) + a \exp(- \hat{q}(\mathbf{x}^\prime_{1:m} \vert h, \X_0))}} \, \cdot \, \frac{\phi(\mathbf{z}_{1:m})}{\Phi(\mathbf{z}_{1:m})}  \\
    \end{bmatrix} + \mathbf{L}^{-1} \mathbf{f}  \quad \in \mathbb{R}^{2m} \\
    & \frac{\partial^2}{\partial \ff \,     \partial \ff^\top} \, S(\mathbf{f}) = \Lambda + \mathbf{L}^{-1} \quad \in \mathbb{R}^{2m \times 2m} \\
    & \Lambda = \begin{bmatrix}
        \cc \cdot \mathrm{diag}\left(\frac{\phi(\mathbf{z}_{1:m})^2}{\Phi(\mathbf{z}_{1:m})^2}  + \frac{\phi(\mathbf{z}_{1:m})}{\Phi(\mathbf{z}_{1:m})} \mathbf{z}_{1:m} \right) & \cc \cdot \mathrm{diag} \left( - \frac{\phi(\mathbf{z}_{1:m})^2}{\Phi(\mathbf{z}_{1:m})^2}  - \frac{\phi(\mathbf{z}_{1:m})}{\Phi(\mathbf{z}_{1:m})} \mathbf{z}_{1:m} \right) \\
        \cc \cdot \mathrm{diag} \, \left( - \frac{\phi(\mathbf{z}_{1:m})^2}{\Phi(\mathbf{z}_{1:m})^2}  - \frac{\phi(\mathbf{z}_{1:m})}{\Phi(\mathbf{z}_{1:m})} \mathbf{z}_{1:m} \right) & \cc \cdot \mathrm{diag}\left( \frac{\phi(\mathbf{z}_{1:m})^2}{\Phi(\mathbf{z}_{1:m})^2}  + \frac{\phi(\mathbf{z}_{1:m})}{\Phi(\mathbf{z}_{1:m})} \mathbf{z}_{1:m} \right) \\
    \end{bmatrix} \quad \in \mathbb{R}^{2m \times 2m} \\
    & \cc =  \left [ \frac{1}{a \exp(- \hat{q}_X(\mathbf{x}_{1:m} \vert h, \X_0)) +  a \exp(- \hat{q}_X(\mathbf{x}^\prime_{1:m} \vert h, \X_0))} \right] \quad  \in \mathbb{R}^{m}
\end{align}

where $\phi$ denotes the probability density function of standard normal distribution. Subsequently, the first and second derivatives of the \textbf{logistic likelihood} function are given by
\begin{align}
 & \frac{\partial}{\partial \ff} \, S(\mathbf{f}) = 
    \begin{bmatrix}
         - \frac{1 / \lambda(\x_{1:m}) \exp( f(\x'_{1:m}) / \lambda(\x'_{1:m}) ) }{\exp(f(\x_{1:m}) / \lambda(\x_{1:m})) + \exp(f(\x'_{1:m}) / \lambda(\x'_{1:m}))} \\
         \frac{1 / \lambda(\x'_{1:m}) \exp( f(\x'_{1:m}) / \lambda(\x'_{1:m}) ) }{\exp(f(\x_{1:m}) / \lambda(\x_{1:m})) + \exp(f(\x'_{1:m}) / \lambda(\x'_{1:m}))}  \\
    \end{bmatrix} + \mathbf{L}^{-1} \mathbf{f}  \quad \in \mathbb{R}^{2m} \\
    & \frac{\partial^2}{\partial \ff \,     \partial \ff^\top} \, S(\mathbf{f}) = \Lambda + \mathbf{L}^{-1} \quad \in \mathbb{R}^{2m \times 2m} \\
    & \Lambda = \begin{bmatrix}
        \frac{\cc}{\lambda^2(\x_{1:m})}  & \frac{- \cc}{\lambda(\x_{1:m}) \lambda(\x'_{1:m})}  \\
         \frac{- \cc}{\lambda(\x_{1:m}) \lambda(\x'_{1:m})} & \frac{\cc}{\lambda^2(\x'_{1:m})} \\
    \end{bmatrix} \quad \in \mathbb{R}^{2m \times 2m} \\
    & \cc =  \left [ \frac{\exp \left( f(\x_{1:m}) / \lambda(\x_{1:m}) + f(\x'_{1:m}) / \lambda(\x'_{1:m}) \right)}{(\exp(f(\x_{1:m}) / \lambda(\x_{1:m})) + \exp(f(\x'_{1:m}) / \lambda(\x'_{1:m})))^2} \right] \quad  \in \mathbb{R}^{m}
\end{align}

The solution of $\frac{\partial}{\partial \ff} S(\mathbf{f}) = 0$ provides $\ff_\mathrm{MAP}$. We use the Newton-Raphson method to obtain $\ff_\mathrm{MAP}$, following~\citet{chu2005preference}. The Laplace method approximates $S(\ff)$ as a Gaussian distribution with mean vector $\ff_\mathrm{MAP}$ and the covariance matrix $(\mathbf{L}^{-1} + \Lambda_\mathrm{MAP})^{-1}$, where $\Lambda_\mathrm{MAP} \triangleq \Lambda \vert_{\ff_\mathrm{MAP}}$. Given $\ff_\mathrm{MAP}$ and $\Lambda_\mathrm{MAP}$, we approximate the marginal likelihood as 
\begin{align}\label{eq:laplace-evidence}
p(\mathcal{D} \vert \theta) \approx \exp(- S(\mathbf{f}_\mathrm{MAP})) \, \mathrm{det}(\mathbf{I} + \mathbf{L} \Lambda_\mathrm{MAP} )^{- 1/2}.
\end{align}
The hyperparameter optimization problem can be formulated as
\begin{align}\label{eq:hyperparam-problem}
    \theta^\ast = \arg \min_{\theta} - \log p(\mathcal{D} \vert \theta), 
\end{align}
with $\theta$ denoting the lengthscale of GP $f$. We apply L-BFGS-B \citep{nocedal1980updating} to obtain $\theta^\ast$.

Subsequently, we run leave-one-out (LOO) \citep{hastie2009elements} to optimize the KDE kernel bandwidth $h$. Note that LOO is a special case of cross-validation, where each fold is of size one. 
The usually computationally expensive technique remains amenable in the context of our study, where human experts typically provide a small number of \emph{anchors}. Given the \emph{anchors} $\X_0$ and bandwidth $h$, we aim to minimize the negative LOO:
\begin{equation}
h^\ast \triangleq \arg \min_{h} - \mathrm{LOO}(h) = - \frac{1}{n} \sum_{\x_0 \in \X_0} \log \hat{q}(\x_0 \vert h, \X_0 \setminus \{\x_0\}).
\end{equation}
For each anchor $\x_0 \in \X_0$, we employ the remaining \emph{anchors} $\X_0 \setminus \{\x_0\}$ to estimate $q(\x_0)$. We then take the negative average of the logarithmic estimator for all \emph{anchors}. We also employ L-BFGS-B to obtain $h^\ast$. 

According to Stone's theorem, the KDE bandwidth $\hat{h}$ derived through cross-validation converges to the optimal $h$ value \citep{stone1974cross}. This optimal bandwidth minimizes the Mean Squared Error (MSE) between the true density $q(\x)$ and the density estimator $q(\x \vert h, \X_0)$ for all $\x$. This theorem leads to the proposition below.
\begin{proposition}
    Suppose that $\sigma_\varepsilon^2(\x)$ is bounded. Let $\hat{\sigma}^2_\varepsilon(\x \vert h)$ denote the kernel variance estimator with bandwidth $h$ and let $\hat{\sigma}^2(\x \vert \hat{h})$ denote the bandwidth chosen by leave-one-out. Then,
    \begin{equation}\label{eq:stone-theorem}
        \frac{ \int (\sigma_\varepsilon^2(\x) - \hat{\sigma}_\varepsilon^2(\x \vert \hat{h}) )^2 \mathrm{d}\x}{\underset{h}{\inf} \int (\sigma_\varepsilon^2(\x) - \hat{\sigma}_\varepsilon^2(\x \vert h))^2 \mathrm{d}\x} \xrightarrow{\text{a.s.}} 1.
    \end{equation}
\end{proposition}
Note that we slightly abuse the notation of the estimator variance to differentiate the bandwidth being utilized. The result follows the \textit{Lipschitz continuous} property of the noise variance.

\subsection{Time complexity of KDE estimator}
For a set of anchors $\X_0$ with $N$ samples and $M$ evaluation inputs $\hat{\x}_1, \dots, \hat{\x}_m \in \mathcal{X}$, the time complexity of the density estimator $\hat{q}$ is $\mathcal{O}(NM)$ \citep{raykar2010fast}. This arises from constructing kernel matrix sized $N \times M$ and the number of operations to obtain the density estimator for $m$ evaluation inputs. Note that this estimator does not hurt the time complexity as GP exhibits time complexity of $\mathcal{O}(N^3)$ \citep{books/lib/RasmussenW06}. 

\section{Details of heteroscedastic Hallucination Believer inference}\label{app:HB}

Given the set of duels $\vv_m < 0$ specified in Appendix \ref{app:likelihoods}, \citet{takeno2023towards} propose to generate a new sample $\tilde{\vv}_m$, called the hallucination drawn from $p(\vv_m \vert \vv_m < 0)$ through Gibbs sampling. Note that the sample path from the skew GP denoted as $p(\mathbf{f}_\ast \vert \mathbf{v}_m < 0)$ for any output vector of the inputs $\mathbf{f}_\ast \triangleq [f(\mathbf{x}_1^\ast), \ldots, f(\mathbf{x}_t^\ast)]^\top$ follows an MVN distribution, implying a regular GP. This characteristic enables a proper posterior computation of $p(\ff_\ast \vert \vv_m < 0, \Tilde{\vv}_m)$, outperforming other approximate inference techniques like Laplace approximation. Nevertheless, we emphasize that the proposed noise distribution can be applied to any approximate inference method.

\begin{algorithm}[h]
\caption{Hallucination Believer (\citep{takeno2023towards}) for posterior approximation and query acquisitions}
\label{algo:HB}
    \begin{algorithmic}[1]
        \STATE \textbf{Input}: Initial dataset $\mathcal{D} = \{\x_k \succ \x^\prime_k\}_{k=1}^m$
        
        \FOR {$t=1,\dots$}
                \STATE $\x_{t} \gets \x_{t-1}$ ~~~ \texttt{//set previous winner as first design of the pair}
            \STATE Draw $\tilde{\mathbf{v}}_{t-1}$ from the posterior $p(\mathbf{v}_{t-1}|\mathbf{v}_{t-1} < \mathbf{0})$ via Gibbs sampling of truncated MVN
            \STATE Sequentially estimate $f$ and $\sigma^2_\varepsilon(\x)$ (Section~\ref{subsec:simplemodel})
            \STATE $\hat{\x}_t \gets \argmax_{\x \in \mathcal{X}} \alpha(\x)$ based on GPs
            $f|\tilde{\mathbf{v}}_{t-1}$ and $\varepsilon(\x) \sim \mathcal{N}(0, \hat{\sigma}^2_\varepsilon(\x))$ 
            \STATE Set the winner as $\x_t$ and the loser as $\x^\prime_t$,
            respectively.
            %\vspace{-.4cm}
            \STATE $\mathcal{D}_t \gets \mathcal{D}_{t-1} \cup (\x_{t} \succ \x^\prime_{t})$
        \ENDFOR
    \end{algorithmic}
\end{algorithm}

\subsection{Posterior predictive distribution}

Following \citet{takeno2023towards}, we formalize the predictive posterior as follows. We first model the joint distribution between the test function $\ff_\ast$ and the hallucination $\vv_m$. We obtain $\vv_m$ by running the Gibbs sampling algorithm for the truncated MVN.
\begin{align}
   &\begin{bmatrix}
        \mathbf{f}_* \nonumber \\
        \vv_m
    \end{bmatrix} \sim \mathcal{N}(\mathbf{0}, \boldsymbol{\Sigma})  \\ &\boldsymbol{\Sigma} \triangleq \mathbf{A} (\mathbf{L} + \mathbf{B}) \mathbf{A}^\top \in \mathbb{R}^{(t + m) \times (t + m)}   \\
    &\mathbf{A} \triangleq \begin{bmatrix}
        \mathbf{I}_n & \mathbf{0} & \mathbf{0} \\
        \mathbf{0} & - \mathbf{I}_m & \mathbf{I}_m  \\
    \end{bmatrix}  \in \mathbb{R}^{(t + m) \times (t + 2m)}  \\
    &\mathbf{B} \triangleq \begin{bmatrix}
        \mathbf{0} & \mathbf{0} \\
        \mathbf{0} & \mathbf{V}_{\text{noise}}
    \end{bmatrix} \in \mathbb{R}^{(t + 2m) \times (t + 2m)}, 
\end{align} 

where $\mathbf{L} = \{l(\x, \x^\prime) \}_{\x, \x^\prime \in \X \cup \X^\ast}$ and $\mathbf{V}_\mathrm{noise} = \mathrm{diag}([a \exp(- \hat{q}(\x \vert h, \X_0))]_{\x \in \X})$. We then rewrite the kernel matrix $\bsigma$ into a block of matrices as follows:
\begin{align}\label{eq:joint-matrix}
    \mathbf{\Sigma} \triangleq \begin{bmatrix}
        \boldsymbol{\Sigma}_{\ast, \ast} & \boldsymbol{\Sigma}_{\ast, \vv_m} \\
        \boldsymbol{\Sigma}_{\vv_m, \ast} & \boldsymbol{\Sigma}_{\vv_m, \vv_m}
    \end{bmatrix}
\end{align}
where $\bsigma_{*,*} \in \R^{t \times t}$, $\bsigma_{*, \vv_m} \in \R^{t \times m}$, $\bsigma_{\vv_m, \vv_m} \in \R^{m \times m}$. For an output vector $\ff_\ast$ of a given predictive inputs $\mathbf{X}^\ast = \{\x^\ast_1, \dots,  \x^\ast_t\}$, the Hallucination Believer predictive posterior $p(\mathbf{f}_\ast \vert \mathbf{v}_m < 0)$ is defined as:
\begin{align}\label{eq:posterior-hb}
 &p(\mathbf{f}_* \vert \mathbf{v}_m < 0, \mathbf{v}_m) = p(\ff_\ast \vert \mathbf{v}_m) \triangleq \mathcal{N}(\ff_\ast \, \vert \, \boldsymbol{\mu}_{* \vert \mathbf{v}_m}, \boldsymbol{\Sigma}_{* \vert \mathbf{v}_m}) \nonumber \\
&\boldsymbol{\mu}_{* \vert \mathbf{v}_m} \triangleq \boldsymbol{\Sigma}_{*, \mathbf{v}_m} \, \boldsymbol{\Sigma}^{-1}_{\mathbf{v}_m, \mathbf{v}_m} \, \mathbf{v}_{m - 1} \\
&\boldsymbol{\Sigma}_{* \vert \mathbf{v}_m} \triangleq  \boldsymbol{\Sigma}_{*, *}  - \boldsymbol{\Sigma}_{*, \mathbf{v}_m} \, \boldsymbol{\Sigma}_{\mathbf{v}_m, \mathbf{v}_m}^{- 1} \, \boldsymbol{\Sigma}_{*, \mathbf{v}_m}^\top
\end{align}
\subsection{Gibbs sampling for truncated MVN}

Here, we provide the Gibbs sampling algorithm for truncated MVN distribution to draw the hallucination $\vv_m$ in the HB inference algorithm. The algorithm utilizes $\bsigma$ from equation \ref{eq:joint-matrix} to update each component of $\vv_m$.  

\begin{algorithm}[h]
\caption{Gibbs sampling for truncated MVN (\citet{takeno2023towards})}
\label{algo:trunc-MVN}
    \begin{algorithmic}[1]
        \STATE \textbf{Input}: $\vv_0 = \mathbf{0}, \bsigma$ (Equation \ref{eq:joint-matrix})
        \STATE Compute $\bsigma^{-1}$
        \FOR {$i=1,\dots$}
           \STATE $\vv_i \leftarrow \vv_{i - 1}$ 
           \FOR{$j=1, \dots$}
                \STATE $\mu_{i, j} \rightarrow [\bsigma^{-1} \vv_i]_j / [\bsigma^{-1}]_{j, j}$
                \STATE Set $\vv_{i, j}$ by sampling from $\mathcal{N}(v_{i, j} \vert \mu_{i, j}, 1 / [\sigma^{-1}]_{j, j})$ with truncation above at 0
            \ENDFOR
        \ENDFOR
    \end{algorithmic}
\end{algorithm}

\section{Details of heteroscedastic Laplace inference}\label{app:detail-LA}

We describe the formulation of Laplace predictive distribution with the heteroscedastic noise as follows. For an output vector $\ff_\ast$ of a given predictive inputs $\mathbf{X}^\ast = \{\x^\ast_1, \dots,  \x^\ast_t\}$, the Laplace predictive posterior $p(\mathbf{f}_\ast \vert \mathbf{v}_m < 0)$ is defined as:
\begin{align}
    &p(\mathbf{f}_\ast \vert \mathbf{v}_m < 0) \triangleq \normal(\ff_\ast \vert \, \bmu_{\ast \vert \vv_m}, \bsigma_{\ast \vert \vv_m}) \nonumber \\ 
    &\bmu_{\ast \vert \vv_m} =  \bsigma_{\ast, \vv_m} \, \bsigma_{\vv_m, \vv_m}^{-1} \, \ff_\mathrm{MAP} \\
    &\bsigma_{\ast \vert \vv_m} = \bsigma_{\ast, \ast} -  \bsigma_{\ast, \vv_m} \, (\bsigma_{\vv_m, \vv_m} + \Lambda_\mathrm{MAP})^{-1}  \bsigma_{\ast, \vv_m}^\top
\end{align}
%
%where $\mathbf{L}_{\ast, \vv_m} \in \R^{t \times 2m} \triangleq l(\y, \x), \mathbf{L}_{\ast, \ast} \in \R^{t \times t} \triangleq l(\y, \y^\prime) \, \forall \x \in \mathbf{X}$ and $\forall \y, \y^\prime \in \X^\ast$. 

where $\bsigma_{\ast, \ast}, \bsigma_{\ast, \vv_m}, \bsigma_{\vv_m, \vv_m}$ follow \ref{eq:joint-matrix} and $\ff_\mathrm{MAP}, \Lambda_\mathrm{MAP}$ follows Appendix \ref{app:hyperparameter-optimization}. Note that the forms of $\ff_\mathrm{MAP}$ and $\Lambda_\mathrm{MAP}$ depend on the likelihood function.

\section{Details of heteroscedastic expectation propagation inference}\label{app:detail-EP}

Following \citep{takeno2023towards}, we apply expectation propagation (EP) directly on $\vv_m \vert \vv_m < 0$. Given $\vv_m \sim \normal(\mathbf{0}, \bsigma_{\vv_m, \vv_m})$, we define the EP approximation as follows:
\begin{align}
    &p(\vv_m \vert \vv_m < 0) \propto \normal(\vv_m \vert \mathbf{0}, \bsigma_{\vv_m, \vv_m}) \prod_{i = 1}^m \mathbb{I}(v_i < 0) \\
    & \approx \normal(\vv_m \vert \mathbf{0}, \bsigma_{\vv_m, \vv_m})  \prod_{i = 1}^m \normal(v_i \vert \tilde{\mu}_i, \tilde{\sigma}_i^2) \\
    & = \normal(\vv_m \vert \mathbf{0}, \bsigma_{\vv_m, \vv_m}) \normal(\vv_m \vert \tilde{\bmu}, \tilde{\bsigma}) \\
    &\propto \normal(\vv_m \vert \hat{\bmu} = \hat{\bsigma} (\tilde{\bsigma}^{-1} \tilde{\bmu} + \bsigma_{\vv_m, \vv_m}^{-1} \mathbf{0}), \hat{\bsigma} = (\bsigma_{\vv_m, \vv_m}^{-1} + \tilde{\bsigma}^{-1})^{-1}),
\end{align}
where $\tilde{\bmu} \in \R^m \triangleq [\tilde{\mu}_i, \dots, \tilde{\mu}_m]^\top$ and $\tilde{\bsigma} \in \R^{m \times m} \triangleq \mathrm{diag}([\tilde{\sigma}^2_i, \dots, \tilde{\sigma}^2_m])$. We first consider the \emph{cavity distribution}
\begin{align}
    \Tilde{p}_{\setminus j}(\vv_m) &= \normal(\vv_m \vert \Bar{\bmu}_{ \setminus j}, \Bar{\bsigma}_{ \setminus j}) \nonumber \\
    &\propto \normal(\vv_m \vert \mathbf{0}, \bsigma_{\vv_m, \vv_m}) \prod_{i = 1, i \neq j}^m \normal(v_i \vert \tilde{\mu}_i, \tilde{\sigma}_i^2),
\end{align}
where
\begin{align}
    &\bar{\mu}_{\setminus j j} = \mean_{p_{\setminus j}}[v_j] = \bar{\sigma}^2_{\setminus j j} (\hat{\mu}_j / \hat{\sigma}^2_j - \tilde{\mu}_j / \tilde{\sigma}^2_j) \\
    &\bar{\sigma}^2_{\setminus j j} = \var_{\Tilde{p}_{\setminus j}}[v_j] = (1 / \hat{\sigma}^2_j - 1 / \tilde{\sigma}^2_j)^{-1}.
\end{align}
Given $\Tilde{p}_{\setminus j}(\vv_m)$, we define \emph{tilted} distribution as
\begin{align}
    \Tilde{p}_{\_ j}(v_j) = \underbrace{\Tilde{p}_{\setminus j}(v_j)}_\text{tilted dist.} \underbrace{\frac{\mathbb{I}(v_j < 0)}{\phi(\zeta)}}_{\text{truncated MVN}},
\end{align}
where $\zeta = - \bar{\mu}_{\setminus j j} / \bar{\sigma}_{\setminus j j}$. The expectation propagation iteratively optimizes $\tilde{\bmu}$ and $\tilde{\bsigma}$ as a part of the global approximation. This is done by minimizing the following Kullback-Leibler (KL) -divergence:
\begin{align}
\mathrm{KL}[\Tilde{p}_{\setminus j}(v_j) \Vert \Tilde{p}_{\_ j}(v_j) \normal(v_j \vert \tilde{\mu}_j, \tilde{\sigma}^2_j) ] = \int \Tilde{p}_{\setminus j}(v_j) \log \frac{ \Tilde{p}_{\setminus j}(v_j) }{ \Tilde{p}_{\_ j}(v_j) \normal(v_j \vert \tilde{\mu}_j, \tilde{\sigma}^2_j) } \mathrm{d}v_j
\end{align}
In practice, the minimization requires updating $\tilde{\mu}_j$ and $\tilde{\sigma}^2_j$ using the following rules:
\begin{align}
    &\tilde{\mu}_j \leftarrow \mean_{\Tilde{p}_{\setminus j}}[v_j] + \left( \zeta + \frac{\phi(\zeta)}{\Phi(\zeta)} \right) \bar{\sigma}_{\setminus j j} \\
    &\tilde{\sigma}^2_j \leftarrow \left( \left( \frac{\zeta \phi(\zeta)}{\Phi(\zeta)}  + \left(\frac{\phi(\zeta)}{\Phi(\zeta)}\right)^2 \right)^{-1} - 1\right) \bar{\sigma}^2_{\setminus j j}
\end{align}

For an output vector $\ff_\ast$ of a given predictive inputs $\mathbf{X}^\ast = \{\x^\ast_1, \dots,  \x^\ast_t\}$, the EP predictive posterior $p(\mathbf{f}_\ast \vert \mathbf{v}_m < 0)$ is defined as:
\begin{align}
   &p(\mathbf{f}_\ast \vert \mathbf{v}_m < 0) \triangleq \normal(\ff_\ast \vert \, \bmu_{\ast \vert \vv_m}, \bsigma_{\ast \vert \vv_m}) \nonumber \\ 
    &\bmu_{\ast \vert \vv_m} =  \bsigma_{\ast, \vv_m} \, (\bsigma_{\vv_m, \vv_m} + \tilde{\bsigma})^{-1} \, \Tilde{\bmu} \\
    &\bsigma_{\ast \vert \vv_m} = \bsigma_{\ast, \ast} -  \bsigma_{\ast, \vv_m} \, (\bsigma_{\vv_m, \vv_m} + \tilde{\bsigma})^{-1}  \bsigma_{\ast, \vv_m}^\top
\end{align}
where $\tilde{\bmu}$ and $\tilde{\bsigma}$ are obtained via EP.

\section{Proofs}\label{app:proofs}

\subsection{Risk-averse one-step Bayes optimal policy of RAEUBO} \label{app:proof-one-step-bayes}

Except for the number of data $m$, we use the same notations as \citet{pmlr-v206-astudillo23a}. Recall that we consider the noise level parameter $\lambda(x) = a \exp( - \hat{q}(x \vert h, \mathbf{X_0}) )$. Note that $\lambda(x)$ is bounded for all $x$, i.e., $\lambda_{\min} \leq \lambda(x) \leq \lambda_{\max}$, where $\lambda_{\min} = \exp(-1) a$ and $\lambda_{\max} = \exp(0) a$. Our objective is to maximize the risk-averse objective $\mathrm{MV(\x)} \triangleq f(\x) - \alpha\lambda(\x)$ for $\alpha >0$. By definition of $\mathrm{MV}$, we have that
\begin{equation}
\mathbb{E}_m [\max_\x \mathbb{E}_m [\mathrm{MV}(\x) \vert \X, r(\X)]] = \mathbb{E}_m[\max_\x \mathbb{E}_m [f(\x) - \alpha \lambda(\x) \vert \X, r(\X)]],
\end{equation}
where $r(\X) = r(\x_1, \x_2) = 1$ if $\x_1$ is preferred over $\x_2$ and $2$ otherwise. $\hat{V}_m^\lambda$ is the risk-averse counterpart of $V^\lambda_n$ defined in [Section A.2] of \citet{pmlr-v206-astudillo23a}. Following \citet{pmlr-v206-astudillo23a}, we first define:
\begin{equation}
    \hat{U}_m^\lambda(X) = \mathbb{E}_m[\mathrm{MV}(\x_{r(\X)})]
\end{equation}
Then, a parallel with [Lemma A.2] from \citet{pmlr-v206-astudillo23a} can be formalized as follows: 

\begin{lemma}\label{lemma:parallel-A2}
    $\hat{U}_n^\lambda(X) \geq \mathrm{RAEUBO}(\X) - \hat{\lambda} C$, where 
    \begin{equation}
    \hat{\lambda} \geq \frac{(\mathrm{MV}(\x_2) - \mathrm{MV}(\x_1)) \lambda_{\min} \, \lambda_{\max}}{\mathrm{MV}(\x_2) \, \lambda_{\max} - \mathrm{MV}(\x_1) \, \lambda_{\min}}, 
    \end{equation}    
and $\mathrm{MV}(\x_2) \geq \mathrm{MV}(\x_1)$ without loss of generality.
\end{lemma}
\textit{Proof:}

 We note that
\begin{equation}
    \mathbb{E}[\mathrm{MV}(\x_{r(\X)}) \vert f(\X)] = \sum_{i = 1}^2 \frac{\exp(f(\x_i) / \lambda(\x_i))}{\sum_{j = 1}^2 \exp(f(\x_j) / \lambda(\x_j))} \mathrm{MV}(\x_i)
\end{equation}
Note that for any $\lambda_i := \lambda(\x_i)$, we can bound [Lemma A.1] in \citet{pmlr-v206-astudillo23a}, i.e, for any $s_i \in \mathbb{R}$ and $i \in \{1,2\}$ by choosing $\hat{\lambda} \geq (s_2 - s_1) \lambda_{\min} \lambda_{\max} / (s_2 \lambda_{\max} - s_1 \lambda_{\min})$. We then find that 
 \begin{equation}
     \sum_{i = 1}^2 \frac{\exp(s_i / \lambda_i)}{\sum_{j = 1}^2 \exp(s_j / \lambda_j)} \geq \sum_{i = 1}^2 \frac{\exp(s_i / \hat{\lambda})}{\sum_{j = 1}^2 \exp(s_j / \hat{\lambda})} \geq \max \{s_1,s_2\} - \hat{\lambda} C
 \end{equation}
where $s_2 > s_1$ without loss of generality. Next, adapting the bound of [Lemma A.1] of \citet{pmlr-v206-astudillo23a}, it holds that
\begin{equation}
  \mathbb{E}[\mathrm{MV}(\x_{r(\X)}) \vert f(\X)]  \geq \max \{\mathrm{MV}(\x_1),\mathrm{MV}(\x_2)\} - \hat{\lambda} C
\end{equation}
Taking expectations $\mathbb{E}_n$ over both sides of the inequality yields the desired result. In our case, this also stems from the fact that our KDE-based noise model $\lambda(\cdot)$ does not depend on the dataset $\mathcal{D}_n$, only on the anchors $\mathbf{\X_0}$.\\

Subsequently, adapting [Lemma A.3] in \citet{pmlr-v206-astudillo23a} provides the following:
\begin{lemma}\label{lemma:parallel-A3}
    $\hat{V}_n^\lambda(\X) \geq \hat{U}_n^\lambda(\X)$ for all $\X$.
\end{lemma}

\textit{Proof:}

Observe that
\begin{align}
    \hat{V}_n^\lambda(\X) &= \mathbb{E}_n[\max_{\x \in \mathcal{X}} \mathbb{E}[\mathrm{MV}(\x) \vert \X, r(\X)]] \\
    &\geq \mathbb{E}_n[\mathbb{E}[\mathrm{MV}(\x_{r(\X)}) \vert \X, r(\X)] ] \\
    &= \mathbb{E}_n[f(\x_{r(\X)}) - \alpha \lambda(\x_{r(\X)})]\\
    &= \hat{U}_n^\lambda(\X)
\end{align}
We can derive a risk-averse version of the Lemmas derived in \citet{pmlr-v206-astudillo23a} with these risk-averse versions of [Theorem 2]. Based on the notations from \citet{pmlr-v206-astudillo23a}, let $\X^* \in \text{argmax}_{\X \in \mathbb{X}^2} \text{ RAEUBO}(\X)$. Subsequently, let $\X^{**} \in \text{argmax}_{\X \in \mathbb{X}^2} \hat{V}^0_n(\X)$. Recall that in our case, $\X=(\x_1, \x_2)$ and let $\X^+(\X^{**}) = (\text{argmax}_{\x \in \mathbb{X}} \mathbb{E}_n[f(\x)| \X^{**},1], \text{argmax}_{\x \in \mathbb{X}} \mathbb{E}_n[f(x)|X^{**},2])$, we have that
\begin{align}
    \hat{V}_n^\lambda(\X^\ast) &\geq \hat{U}_n^\lambda(\X^\ast) \\
    &\geq \hat{U}_n^0(\X^\ast) - \hat{\lambda} C \\
    &= \mathrm{RAEUBO}(\X^\ast) - \hat{\lambda} C \\
    &\geq \mathrm{RAEUBO}(\X^{+}(\X^{\ast \ast})) - \hat{\lambda} C \\
    & \geq \hat{V}_n^0(\X^{\ast \ast}) - \hat{\lambda} C\\&= \max_{\X \in \mathbb{\X}^2} \hat{V}_n^0(\X) - \hat{\lambda} C
\end{align}
The first line follows from Lemma \ref{lemma:parallel-A3}. The second line follows from adapted Lemma \ref{lemma:parallel-A2}. The third line follows from the definition of $\hat{U}^0_n$. The fourth line follows from the definition of $\X^*$. The fifth line can be obtained as in the proof of [Theorem 1] from \citet{pmlr-v206-astudillo23a} (noise-free setting). Finally, the last line follows from the definition
of $\X^{**}$.

\subsection{Consistency analysis of the KDE-based
model of user epistemic uncertainty}\label{app:proof-of-risk}

\subsubsection{Risk analysis}

The proof begins by providing the definitions of \textit{Hölder class} and the \textit{kernel of order} $\ell$, which are used to construct the necessary assumptions. 

We first introduce \emph{multi-index} notation as follows:
\begin{equation*}
    \vert \s \vert = s_1 + \dots + s_d,% \quad \s! = s_1! \dots s_d!,
    \quad \x^s = x_1^{s_1} \dots x_d^{s_d} 
\end{equation*}
with $\s \in \mathbb{N}_0^d, \x \in \R^d$, and $\vert . \vert$ denotes the magnitude of $\s$. Using \emph{multi-index} notation, the derivative of a function $f:\R^d \rightarrow \R$ is denoted by
\begin{equation*}
    D^{\vert \s \vert} f = \frac{\partial^{\vert \s \vert}f}{\partial x_1^{s_1} \dots \partial x_d^{s_d}}
\end{equation*}

\begin{definition}\label{definition:Hölder-class}
    Assume that $\mathcal{X} \subseteq \mathbb{R}^d$ and let $\beta, L > 0$. \textit{The Hölder class} $\Sigma(\beta, L)$ on $\mathcal{X}$ is defined as the set of $\vert \s \vert = \beta - 1$ times differentiable functions $f: \mathcal{X} \rightarrow \mathbb{R}$ whose derivative $D^{\vert \s \vert}f$ satisfies
    \begin{equation}
        \vert D^{\vert \s \vert} f(\x) - D^{\vert \s \vert} f(\y) \vert \leq L \Vert \x - \y \Vert \quad \forall \x, \y \in \mathcal{X}
    \end{equation}
    with $\Vert . \Vert$ denotes the metric on $\mathcal{X}$.
\end{definition}

\begin{definition}\label{definition:kernel-order-l}
    Let $\ell \geq 1$ be an integer. We say that $k: \mathbb{R} \rightarrow \mathbb{R}$ is a \textit{kernel of order $\ell$} if the functions $r \mapsto r^j k(r), j = 0, \dots, \ell$ are integrable and satisfy
    \begin{equation}
        \int k(r) dr = 1, \quad \int r^j k(r) dr = 0 \quad j=1, \dots, \ell
    \end{equation}
\end{definition}

It is also important to introduce the notion of \emph{Lipschitz continuity} as the foundation for our proofs.

\begin{definition}
 A function $f: \mathcal{X} \subseteq \R^d \rightarrow \R$ is Lipschitz-continuous if there exists $K > 0$ such that, for all $\x, \y \in \mathcal{X}$
\begin{equation}
    \vert f(\x) - f(\y) \vert \leq K \Vert \x - \y \Vert
\end{equation}
with $K$ and $\Vert . \Vert$ denote the Lipschitz constant and metric on $\mathcal{X}$, respectively.
\end{definition}

Subsequently, we mention all the propositions responsible for developing Lemma \ref{lemma:mse-p}. Later on, we also employ these propositions to prove \Cref{prop:concentration-inequality}. The propositions provide the bound for the variance and the bias of the KDE, respectively.

\begin{proposition}\label{proposition:variance}
(\citet{sen2020introductionNP}) 
Suppose that the density $q(\x)$ satisfies $q(\x) \leq q_\mathrm{max} < \infty$ for all $\x \in  \mathbb{R}^d$. Let $k : \mathbb{R} \rightarrow \mathbb{R}$ be the kernel such that

\begin{center}
    $\int k^2( r ) dr \leq \infty$
\end{center}
Then, for any $\x \in \R^d, \; h > 0, \;$ and $n \geq 1$ we have
\begin{equation}
\mathbb{V}\left[ \hat{q}(\x \vert h, \X_0) \right] \leq \frac{ c_1}{n h^d}
\end{equation}
with $c_1 \triangleq q_\mathrm{max} \int k^2( r ) dr$.
\end{proposition}

\begin{proposition}\label{proposition:bias}
    (\citet{sen2020introductionNP})
    Assume $q \in \mathcal{P}(\beta, L)$ and let $k$ be a kernel of order $\ell = \lfloor \beta \rfloor$. Then for any $\x \in \mathbb{R}^d, h > 0,$ and $n \geq 1$ we have 
    \begin{equation}
        \vert \mathbb{E}[\hat{q}(\x \vert h, \X_0)] - q(\x) \vert \leq c_2 h^\beta
    \end{equation}
    where $c_2 \triangleq \frac{L}{\ell !} \int \vert r \vert^\beta  k( r )  dr$
\end{proposition}

The proof of Proposition \ref{prop:mse} relies on the following lemma, which provides the bound of the MSE between the density estimator and the true probability density function whenever Assumption \ref{assumption:risk-analysis} holds.

\begin{lemma}\label{lemma:mse-p}
    (\citet{sen2020introductionNP}) Suppose that Assumption \ref{assumption:risk-analysis} holds. Fix $\alpha > 0$ and take $h = \alpha n^{-1/(2\beta + d)}$. Then, for any $\x$ and $n \geq 1$, the estimated variance $\hat{q}(\x \vert h, \X_0)$ satisfies
    \begin{equation}
        \underset{q \in \mathcal{P}(\beta, L)}{\sup} \mathbb{E}_{\mathbf{X_0}}[(\hat{q}(\x \vert h, \X_0) - q(\x))^2] \leq c_3 n^{- \frac{2 \beta}{2 \beta + d}}
    \end{equation}
    where $c_3 > 0$ is a constant depending only on $\beta, \alpha$ and on the kernel bandwidth $h$.
\end{lemma}

~\Cref{lemma:mse-p} obtains the bound by decomposing MSE as the addition of variance and bias squared. Subsequently, it leverages ~\Cref{proposition:variance} and ~\Cref{proposition:bias} to obtain the bound of variance and bias, respectively. Equipped with~\Cref{lemma:mse-p}, we can now state and prove~\Cref{prop:mse}.

\riskanalysisproposition*

\textbf{Proof:}

By invoking the definition of the variance of the noise distribution, we derive:
\begin{equation}\label{eq:variance-definition}
    (\hat{\sigma}_\varepsilon^2(\x) - \sigma_\varepsilon^2(\x))^2 = \left( \vert \hat{\sigma}_\varepsilon^2(\x) - \sigma_\varepsilon^2(\x) \vert \right)^2 = a^2 \left( \vert  \exp(- \hat{q}(\x \vert h, \X_0)) - \exp(- q(\x))  \vert \right)^2.
\end{equation}
%
%Note that both $p(\x)$ and $\hat{p}(\x \vert h, \X_0)$ denote quantities bounded in $[0,1]$. % $0 \leq p(\x), \, p(\x \vert h, \X_0) \leq 1$ for all $\x$.
Define the map $g :s \mapsto \exp(- s)$ for $s \in [0, 1]$. By the Mean Value Theorem, $g$ is \emph{Lipschitz-continuous} with Lipschitz constant $K = 1$. Furthermore, $q(\x)$ and $\hat{q}(\x \vert h, \X_0)$ are quantities bounded in $[0,1]$ for all $\x$.  Then, the following bound holds:
\begin{equation}\label{eq:lipschitz-p}
    \vert  \exp(- \hat{q}(\x \vert h, \X_0)) - \exp(- q(\x))  \vert \leq \vert \hat{q}(\x \vert h, \X_0) - q(\x) \vert
\end{equation}
We then apply Lemma \ref{lemma:mse-p} to obtain:
\begin{align}
    \underset{q \in \mathcal{P}(\beta, L)}{\sup} \mathbb{E}_{\mathbf{X_0}}[a^2 \left( \vert  \exp(- \hat{q}(\x \vert h, \X_0)) - \exp(- q(\x))  \vert \right)^2]  
    &\leq a^2 \underset{q \in \mathcal{P}(\beta, L)}{\sup}\mathbb{E}_{\mathbf{X_0}}[(\hat{q}(\x \vert h, \X_0) - q(\x))^2] \nonumber \\
    &\leq a^2 c_3 n^{- \frac{2 \beta}{2 \beta + d}}
\end{align}
The choice of $h$ is based on the optimum bandwidth of the regular KDE \citep{sen2020introductionNP}.

\subsection{Concentration analysis}\label{app:proof-of-concentration}
This analysis requires an assumption that depends on $\omega-$covering number. As a starting point, we define $\omega-$covering as follows.

\begin{definition}
Let $(\mathcal{X}, \Vert . \Vert)$ be a metric space and $\mathcal{S} \subset \mathcal{X}$. $\{\x_1, \dots, \x_n \} \in \mathcal{X}^n$ is an $\omega-$covering of $\mathcal{S}$ if $\, \forall \y \in \mathcal{S}$, $\exists i$ such that $\Vert 
\y - \x_i \Vert \leq \omega$  
\end{definition}

Based on the definition above, $\omega-$covering number tells the number of $\omega$-balls to cover a given space $\mathcal{S}$ by allowing the overlaps between the balls. The formal definition of $\omega-$covering number is provided below.

\begin{definition}
(covering number) $N(\mathcal{S}, \Vert . \Vert, \omega)  = \min \{n: \exists \, \omega-\text{covering over} \, \mathcal{S} \, \text{of size} \, n   \}$
\end{definition}

Here, we additionally assume the kernel $k$ belongs to a bounded measurable VC class, as described below.

\begin{assumption}\label{assumption:concentration-analysis}
The KDE kernel $k$ belongs to a collection of measurable functions $\mathcal{F}_h = \left\{\z \mapsto k\left( \frac{\Vert \x - \z \Vert}{h} \right), \x \in \mathcal{X} \subseteq \R^d, h > 0 \right\}$, with $\mathcal{F}_h$ satisfies
\begin{equation}
    \underset{\mathbb{P}}{\sup} \, N \left(\mathcal{F}_h, L_2(\mathbb{P}), \omega \Vert F \Vert_{2} \right) \leq \left(\frac{m}{\omega}\right)^v
\end{equation}
where $N \left(\mathcal{F}_h, L_2(\mathbb{P}), \omega \Vert F \Vert_{2} \right)$ denotes the $\omega-$covering number of metric space $(\mathcal{F}_h, L_2(\mathbb{P}))$, $F$ is the envelope function of $\mathcal{F}$, the constants $m, v > 0$ are the VC characteristics of $\mathcal{F}_h$, and the supremum is taken over the set of all probability measures on $\R^d$.
\end{assumption}

\Cref{prop:concentration-inequality} is built upon the following Lemmas. The first lemma is the renowned \emph{Bernstein's inequality}. The second lemma tells the appropriate $\epsilon$ value to guarantee the probability of KDE exponentially deviating from its expectation depending on the number of \emph{anchors} as well as the dimensionality of the data \citep{gine2002rates}. 

\begin{lemma}\label{lemma:bernstein}
(\citet{bernstein1924modification})
Suppose that $W_1, \dots, W_n$ are i.i.d. random variables with mean $\mu$, $\mathbb{V}[W_i] \leq \sigma^2$, and $\vert W_i \vert \leq b$. Then for any $\epsilon > 0$, the following inequality holds.
\begin{equation}
    \mathbb{P}\left( \vert \Bar{W} - \mu \vert > \epsilon \right) \leq 2 \exp \left(-  \frac{n \epsilon^2}{ 2 \sigma^2 + 2 b \epsilon / 3} \right)
\end{equation}
with $\Bar{W}$ denotes the sample mean.
\end{lemma}

\begin{lemma}\label{lemma:ginne-guillou}
    (~\citet{gine2002rates})
    Suppose the kernel $k$ satisfies Assumption \ref{assumption:concentration-analysis}. Given a fixed $h > 0$, there exists constants $c_3, c_4 > 0$, s.t. $\forall \epsilon > 0$ and all large $n$,
    \begin{equation}\label{eq:ginne-gillou}
        \mathbb{P} \left(\underset{\x \in \mathcal{X}}{\sup} \vert  \hat{q}(\x \vert h, \X_0) - \mathbb{E}[\hat{q}(\x \vert h, \X_0)] \vert >  \epsilon \right) < c_3 \exp\left( - c_4 n h^d \epsilon^2 \right)
    \end{equation}
    Let $h_n \rightarrow 0$ as $n \rightarrow \infty$ in such a way that $\frac{n h_n^d}{\vert \log h_n^d \vert} \rightarrow \infty$. Let 
    \begin{equation}\label{eq:asymptotic-epsilon}
        \epsilon_n \geq \sqrt{\frac{\vert \log h_n \vert}{n h_n^d}}
    \end{equation}
\end{lemma}
Then for all large $n$,~\Cref{eq:ginne-gillou} holds with $h$ and $\epsilon$ are replaced by $h_n$ and $\epsilon_n$, respectively.

We are now ready to prove \Cref{prop:concentration-inequality}.

\concentrationanalysisproposition*

\textbf{Proof:}

By invoking the definition of the noise variance and applying the triangle inequality, we derive the following bounds:
\begin{align}\label{eq:triangle-inequality}
    &\vert \hat{\sigma}^2_\varepsilon(\x) - \sigma_\varepsilon^2(\x) \vert = \vert a \exp(- \hat{q}(\x \vert h, \X_0)) - a \exp(- q(\x)) \vert  \nonumber \\ 
    &\leq \vert a \exp(-\hat{q}(\x \vert h, \X_0)) -  a \exp(- \mathbb{E}[\hat{q}(\x \vert h, \X_0)]) \vert + \vert a \exp(- \mathbb{E}[\hat{q}(\x \vert h, \X_0)]) - a \exp(- q(\x)) \vert \nonumber \\
    &\leq a \left( \vert \hat{q}(\x \vert h, \X_0) -  \mathbb{E}[\hat{q}(\x \vert h, \X_0)] \vert + \vert \mathbb{E}[\hat{q}(\x \vert h, \X_0)] - q(\x) \vert \right)
\end{align}
%
%the conditions for Bernstein inequality
The last inequality follows the \textit{Lipschitz continuous} property as in the proof of Proposition \ref{prop:mse}. Next, we bound the second term using Proposition \ref{proposition:bias}, yielding $a \vert \mathbb{E}[\hat{q}(\x \vert h, \X_0)] - q(\x) \vert \leq a \, c_2 h^\beta$. Subsequently, we address the first term of~\Cref{eq:triangle-inequality}. Define that $\hat{q}(\x \vert h, \X_0) \triangleq \frac{1}{n} \sum_{i = 1}^n G^{(i)}$ with the random variable $G^{(i)} \triangleq \frac{1}{h^d} k\left(\frac{\Vert \mathbf{X}_0^{(i)} - \x \Vert_2}{h}\right)$. Notably, $\vert G^{(i)} \vert \leq \frac{b_1}{h^d}$ where $b_1 = k(\mathbf{0})$. Proposition \ref{proposition:variance} provides $\mathbb{V}[G^{(i)}] \leq \frac{c_1}{h^d}$. By applying \textit{Bernstein's inequality}, we obtain the following bound:
\begin{align}
    \mathbb{P} \left(\vert \hat{q}(\x \vert h, \X_0) - \mathbb{E}[\hat{q}(\x \vert h, \X_0)] \vert > \epsilon \right) &\leq 2 \exp\left( - \frac{n \epsilon^2}{2 c_1 h^{-d} + 2 b_1 h^{-d} \epsilon / 3} \right) \nonumber \\
    &\leq 2 \exp\left( - \frac{n h^d \epsilon^2}{4 c_1} \right) 
\end{align}
whenever $\epsilon \leq 3 c_1$ and $b_1 = 1$. If we choose $\epsilon = \sqrt{4 c_1 \log(2 / \delta) / n h^d}$, then the following bound holds.
\begin{equation}\label{eq:density-bernstein-2}
    \mathbb{P}\left( \vert \hat{q}(\x \vert h, \X_0) - \mathbb{E}[\hat{q}(\x \vert h, \X_0)] \vert > \sqrt{\frac{4 c_1}{n h^d} \log \frac{2}{\delta}} \right) \leq \delta
\end{equation}
Applying~\Cref{eq:density-bernstein-2} to~\Cref{eq:triangle-inequality} results in~\Cref{eq:variance-bernstein-inequality}. To prove the second result, we can derive
\begin{align}
&\sup_{\x \in \mathcal{X}}
\left|
\hat{\sigma}^2_\varepsilon(\x)-\sigma_\varepsilon^2(\x)
\right|
=
\sup_{\x \in \mathcal{X}}
\left|
a\exp(-\hat q(\x \vert h,\X_0)) - a\exp(-q(\x))
\right|
\nonumber \\
&\leq
a\sup_{\x \in \mathcal{X}}
\left|
\exp(-\hat q(\x \vert h,\X_0))
-
\exp(-\mathbb{E}[\hat q(\x \vert h,\X_0)])
\right|
\nonumber \\
&\quad
+
a\sup_{\x \in \mathcal{X}}
\left|
\exp(-\mathbb{E}[\hat q(\x \vert h,\X_0)])
-
\exp(-q(\x))
\right|
\nonumber \\
&\leq
a\left(
\sup_{\x \in \mathcal{X}}
\left|
\hat q(\x \vert h,\X_0)-\mathbb{E}[\hat q(\x \vert h,\X_0)]
\right|
+
\sup_{\x \in \mathcal{X}}
\left|
\mathbb{E}[\hat q(\x \vert h,\X_0)]-q(\x)
\right|
\right)
\nonumber \\
&\leq
a\left(
\sup_{\x \in \mathcal{X}}
\left|
\hat q(\x \vert h,\X_0)-\mathbb{E}[\hat q(\x \vert h,\X_0)]
\right|
+
c_2 h^\beta
\right).
 \label{eq:triangle-ineq-infinity}
\end{align}

The third inequality follows the \textit{Lipschitz continuous} property as in the proof of Proposition \ref{prop:mse}. By applying Lemma \ref{lemma:ginne-guillou} into~\Cref{eq:triangle-ineq-infinity} and choose $\epsilon_n = \sqrt{\frac{1}{c_4 n h_n^d} \log \left(\frac{c_3}{\delta}\right)}$, such that $\frac{1}{c_4} \log \left( \frac{c_3}{\delta} \right) \geq \vert \log h_n \vert$ where $c_3, c_4, \epsilon_n$ and $h_n$ satisfy Lemma \ref{lemma:ginne-guillou}, then the following bound holds
\begin{equation}\label{eq:density-gine-guillou}
\mathbb{P}\left( \underset{\x \in \mathcal{X}}{\sup} \, \vert \hat{q}(\x \vert h, \X_0) - \mathbb{E}[\hat{q}(\x \vert h, \X_0)] \vert > \sqrt{\frac{1}{c_4 n h_n^d} \log \frac{c_3}{\delta}} \right) < \delta
\end{equation}
Applying~\Cref{eq:density-gine-guillou} to~\Cref{eq:triangle-ineq-infinity} results in~\Cref{eq:variance-gine-inequality}.

\section{Additional experiments}\label{app:addexp}

\subsection{Noise model's consistency experiments}\label{app:nmc}

\begin{figure}[htbp]
    \centering
     \resizebox{\textwidth}{!}{%
    \includegraphics[width=0.9 \linewidth]{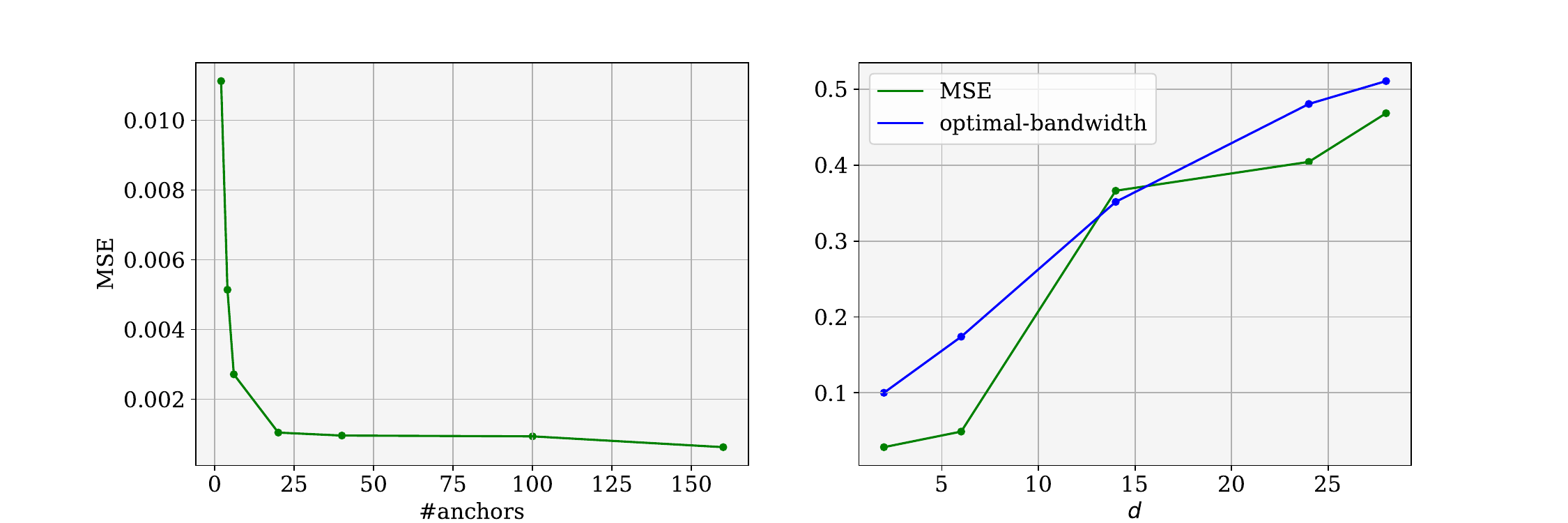}
    }
    \caption{Noise model estimator's consistency results. Left: the estimator's MSE w.r.t. the number of anchors. Right: the estimator's MSE w.r.t. the data dimensionality. Both results are consistent with \Cref{prop:mse}.}
    \label{fig:consistency-results}
\end{figure}

This experiment aims to validate the consistency analysis provided in \Cref{subsec:consistency-analysis}, particularly on Proposition \ref{prop:mse}. The results are shown in Figure \ref{fig:consistency-results}. First, we assess the impact of the number of anchors on the estimator's MSE. This experiment considers a univariate normal distribution with mean $0.5$ and variance $20$ as the ground truth density $q$. The MSE is computed from 500 data samples. The results confirm that, as predicted by Proposition \ref{prop:mse}, the MSE decreases as the number of anchors increases. Subsequently, we examine the effect of data dimensionality $d$ on the estimator's MSE. In this setting, we consider uncorrelated multivariate normal distribution with zero means, fixing the number of anchors at 50 for all scenarios. The results show that the MSE increases with higher data dimensionality, again consistent with Proposition \ref{prop:mse}. Moreover, the optimal kernel bandwidth increases as the dimensionality $d$ grows. 

\subsection{Ranking Plot}

Another way to assess the superiority of our proposed AFs over risk-neutral AFs is to display the average rank as a function of the PBO trial iterations, instead of the achieved reward value as done in Figure~\ref{fig:synthetic-results}.
For each PBO trial, at every iteration $t$, we compute the best value found for each given AF, and rank these values. We then compute the average ranking across all trials. The ranking index starts from $0$. A lower average ranking indicates a better best value found. The results are presented in Figure~\ref{fig:Ranking-results}. Our proposed AFs consistently outperform the baseline in risk-averse settings, except for the ANPEI acquisition function. Conversely, in the risk-neutral scenario, the baseline generally performs better than our methods, except for the Hartmann3 problem, where the RAHBO acquisition function achieves the best ranking.

\begin{figure}[htbp]
    \centering
     \resizebox{ \textwidth}{!}{%
    \includegraphics[width=1\linewidth]{aistats2025/aistats_figure/exp3.pdf}
    }
    \caption{The mean ranking plots of the main experiment presented in Figure \ref{fig:synthetic-results}. The top row depicts the rankings under the risk-averse setting, while the bottom row corresponds to the risk-neutral scenario.}
    \label{fig:Ranking-results}
\end{figure}

\subsection{Computation time}
We compare the computational cost of our method and the baselines across acquisition functions, reporting the mean and standard deviation of the per-iteration runtime. As expected, our method is somewhat slower because it additionally estimates local noise levels, but the overhead remains manageable. Higher-dimensional tasks such as Hartmann4D and Sushi increase runtime for all methods. EUBO and RAEUBO are the most expensive because they jointly optimize duel pairs. Results are reported in Table~\ref{tab:computation_time}.

\begin{table}[H]
\centering
\caption{\small Average computation time (in seconds) and standard deviation per iteration for different acquisition functions.}
\label{tab:computation_time}
\begin{tabular}{c | c c c c c c}
\hline
\textbf{Problem / AF} & UCB & EI & EUBO & RAHBO & ANPEI & RAEUBO \\ \hline
HartMann3 & $25.3 \pm 2.9 $ & $ 26.4 \pm 3.5 $ & $136.8 \pm 20.2$ & $52.4 \pm 6.4$ & $22.8 \pm 2.8$ & $315.1 \pm 29.1$ \\ 
HartMann4 & $37.9 \pm 2.5$ & $ 35.6 \pm 13.5 $ & $213.3 \pm 40.6$ & $57.6 \pm 7.05$ & $33.0 \pm 7.3$ & $ 562.4 \pm 64.3$ \\ 
Sushi  & $38.2 \pm 2.5$ & $33.8 \pm 8.2$ & $231.3 \pm 69.9$ & $45.2 \pm 6.7$ & $34.9 \pm 4.4$ & $539.3 \pm 79.1$ \\ \hline
\end{tabular}
\label{table:ct}
\end{table}

\begin{figure}[h!]
    \centering
    \includegraphics[width=0.85\linewidth]{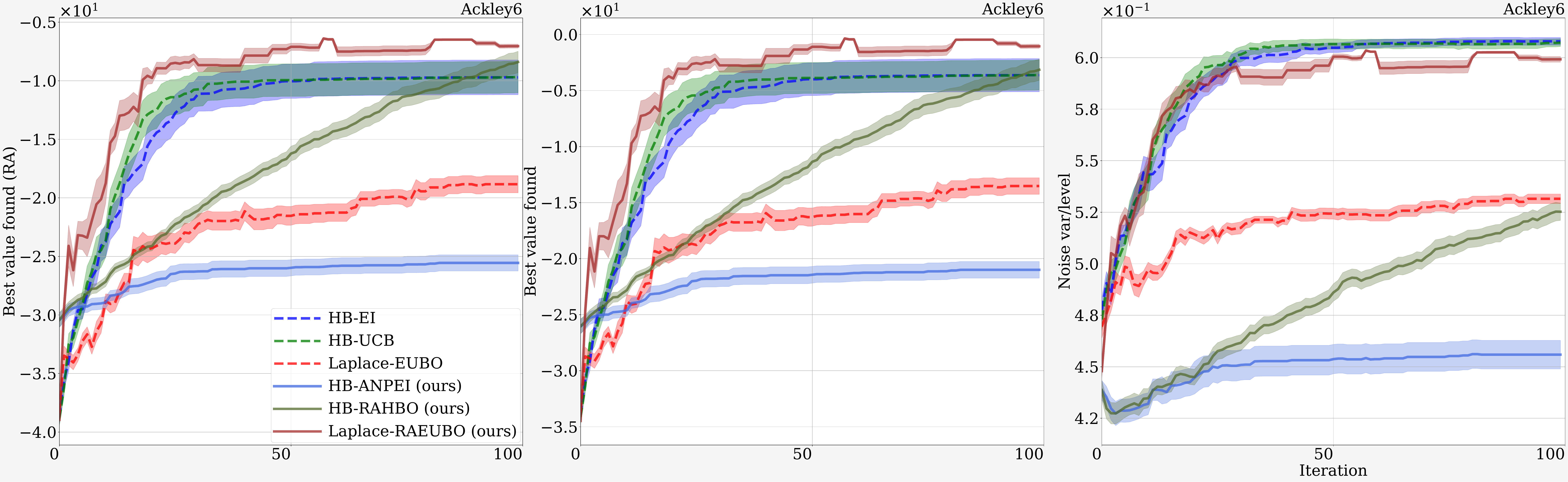}
    \caption{Results on the Ackley 6D benchmark.
    We report performance over 100 PBO iterations, averaged over 20 repetitions.
    (\textbf{Left}) Risk-adjusted best value found.
    (\textbf{Middle}) Conventional best value found.
    (\textbf{Right}) Queried noise variance/level over time.
    Laplace-RAEUBO remains competitive in this higher-dimensional setting, while the KDE-based risk-aware methods degrade more substantially, in line with the known difficulty of KDE in higher dimensions.}
    \label{fig:high-dim-results}
\end{figure}

\begin{figure}[h!]
    \centering
    \includegraphics[width=0.85\linewidth]{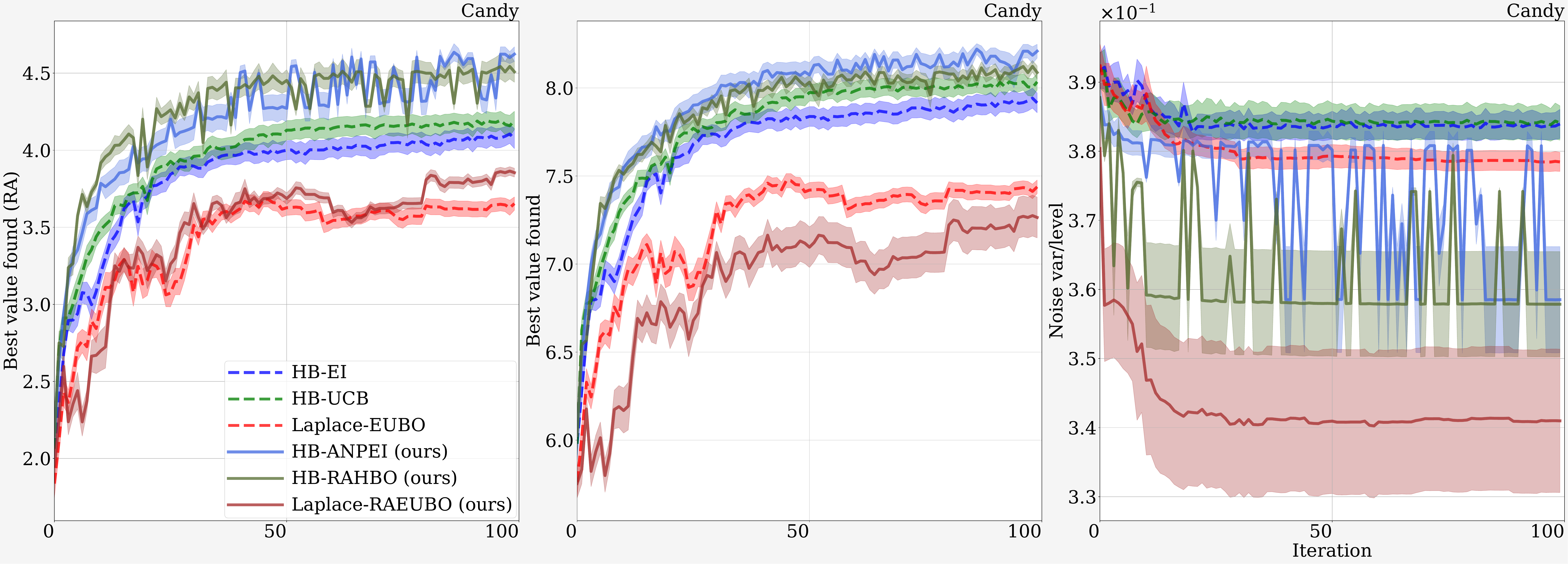}
    \caption{Results on the Candy benchmark.
    We report performance after 100 PBO iterations on the real-world Candy task, averaged over 20 repetitions.
    (\textbf{Left}) Risk-adjusted best value.
    (\textbf{Middle}) Conventional best value found.
    (\textbf{Right}) Estimated queried noise variance/level.
    \textbf{Risk-aware methods generally achieve stronger risk-adjusted performance while remaining competitive on the standard best-value metric, selecting less noisy comparisons than their risk-neutral counterparts.}}
    \label{fig:real-world-results}
\end{figure}

\begin{figure}[h!]
    \centering
    \includegraphics[width=0.85\linewidth]{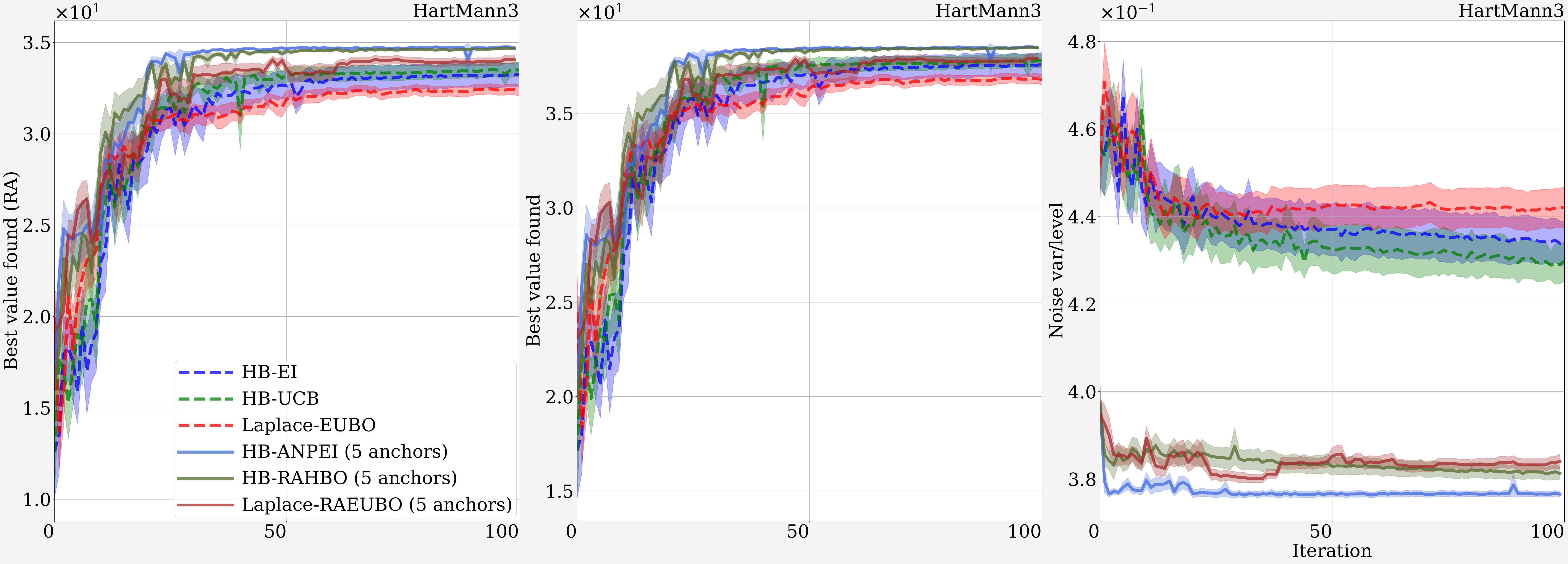}
    \caption{(\textbf{Left}) Risk-averse best value found, (\textbf{Middle}) overall best value found, and (\textbf{Right}) noise variance/level plot for PBO on the Hartmann 3D task after 100 iterations. The results are averaged over 20 repetitions.}
    \label{fig:low-anchors-results}
\end{figure}

\section{Further implementation and experiment details}\label{app:id}

We provide the training details of our experiments to ensure reproducibility and transparency. We set the hyperparameter $a$ defined in Equations \ref{eq:general-noise} and \ref{eq:general-logistic-noise} to $1.0$. We run 30 PBO trials for 150 rounds, each time with a different initial seed. Regarding our method's hyperparameters, the lengthscale $\theta$ is being optimized within $[0.1, 1]$ and the KDE bandwidth $h$ within $[1.0, 2.0]$. The optimal hyperparameters are selected every 10 rounds to adjust to the evolving model complexity as more data becomes available. In terms of acquisition function, we set $\gamma, \alpha=10.0$ and $\eta= 2.0$, following the standard choice in \citet{makarova2021risk}. In the case of the ANPEI acquisition strategy, its closed form is computed:
 \begin{equation}
     \small \alpha_{\mathrm{ANPEI}} = (\mu_{\ast \vert \vv_m}(\x) - \mu_{\ast \vert \vv_m}(\x^\ast)) \Phi \left( \frac{\mu_{\ast \vert \vv_m}(\x) - \mu_{\ast \vert \vv_m}(\x^\ast)}{\sigma^2_{\ast \vert \vv_m}(\x)} \right)  + \sigma_{\ast \vert \vv_m}(\x) \phi\left( \frac{\mu_{\ast \vert \vv_m}(\x) - \mu_{\ast \vert \vv_m}(\x^\ast)}{\sigma^2_{\ast \vert \vv_m}(\x)} \right) -  \gamma \sigma^2_\varepsilon(\mathbf{x}),
 \end{equation}
with $\Phi$ and $\phi$ denote the cumulative and probability density function, respectively. Finally, to kickstart the GP surrogate, each PBO trial is initialized with 8 duels (16 data points) through Sobol sampling.

\end{document}